\documentclass[10pt,twocolumn,letterpaper]{article}

\usepackage[pagenumbers]{iccv} 

\definecolor{iccvblue}{rgb}{0.21,0.49,0.74}
\usepackage[pagebackref,breaklinks,colorlinks,allcolors=iccvblue]{hyperref}

\title{TomatoScanner: phenotyping tomato fruit based on only RGB image}

\author{
Xiaobei Zhao$^{1}$ \quad Xiangrong Zeng$^{1}$ \quad Yihang Ma$^{1}$ \quad Pengjin Tang$^{1}$ \quad Xiang Li$^{1*}$ \\
\small $^{1}$China Agricultural University, Beijing, China \\
\small {\tt 15801175735@163.com, zxr\_zengxiangrong@163.com, yihangma26@gmail.com,} \\
\small {\tt jwxl040720@outlook.com, cqlixiang@cau.edu.cn}
}

\begin{document}
\maketitle
\begin{abstract}
In tomato greenhouse, phenotypic measurement is meaningful for researchers and farmers to monitor crop growth, thereby precisely control environmental conditions in time, leading to better quality and higher yield. 
Traditional phenotyping mainly relies on manual measurement, which is accurate but inefficient, more importantly, endangering the health and safety of people. 
Several studies have explored computer vision-based methods to replace manual phenotyping. 
However, the 2D-based need extra calibration, or cause destruction to fruit, or can only measure limited and meaningless traits. 
The 3D-based need extra depth camera, which is expensive and unacceptable for most farmers. 
In this paper, we propose a non-contact tomato fruit phenotyping method, titled TomatoScanner, 
where RGB image is all you need for input. 
First, pixel feature is extracted by instance segmentation of our proposed EdgeYOLO with preprocessing of individual separation and pose correction. 
Second, depth feature is extracted by depth estimation of Depth Pro. 
Third, pixel and depth feature are fused to output phenotype results in reality. 
We establish self-built Tomato Phenotype Dataset to test TomatoScanner, which achieves excellent phenotyping on width, height, vertical area and volume, 
with median relative error of 5.63\%, 7.03\%, -0.64\% and 37.06\%, respectively. 
We propose and add three innovative modules - EdgeAttention, EdgeLoss and EdgeBoost - into EdgeYOLO, to enhance the segmentation accuracy on edge portion. 
Precision and mean Edge Error greatly improve from 0.943 and 5.641\% to 0.986 and 2.963\%, respectively. 
Meanwhile, EdgeYOLO keeps lightweight and efficient, with 48.7 M weights size and 76.34 FPS. 
Codes and datasets: 
\href{https://github.com/AlexTraveling/TomatoScanner}{https://github.com/AlexTraveling/TomatoScanner}.
\end{abstract}


\section{Introduction}
\label{sec:intro}

\begin{figure}[t]
\centering
    \begin{subfigure}{0.32\linewidth}
        \includegraphics[width=1.0\linewidth]{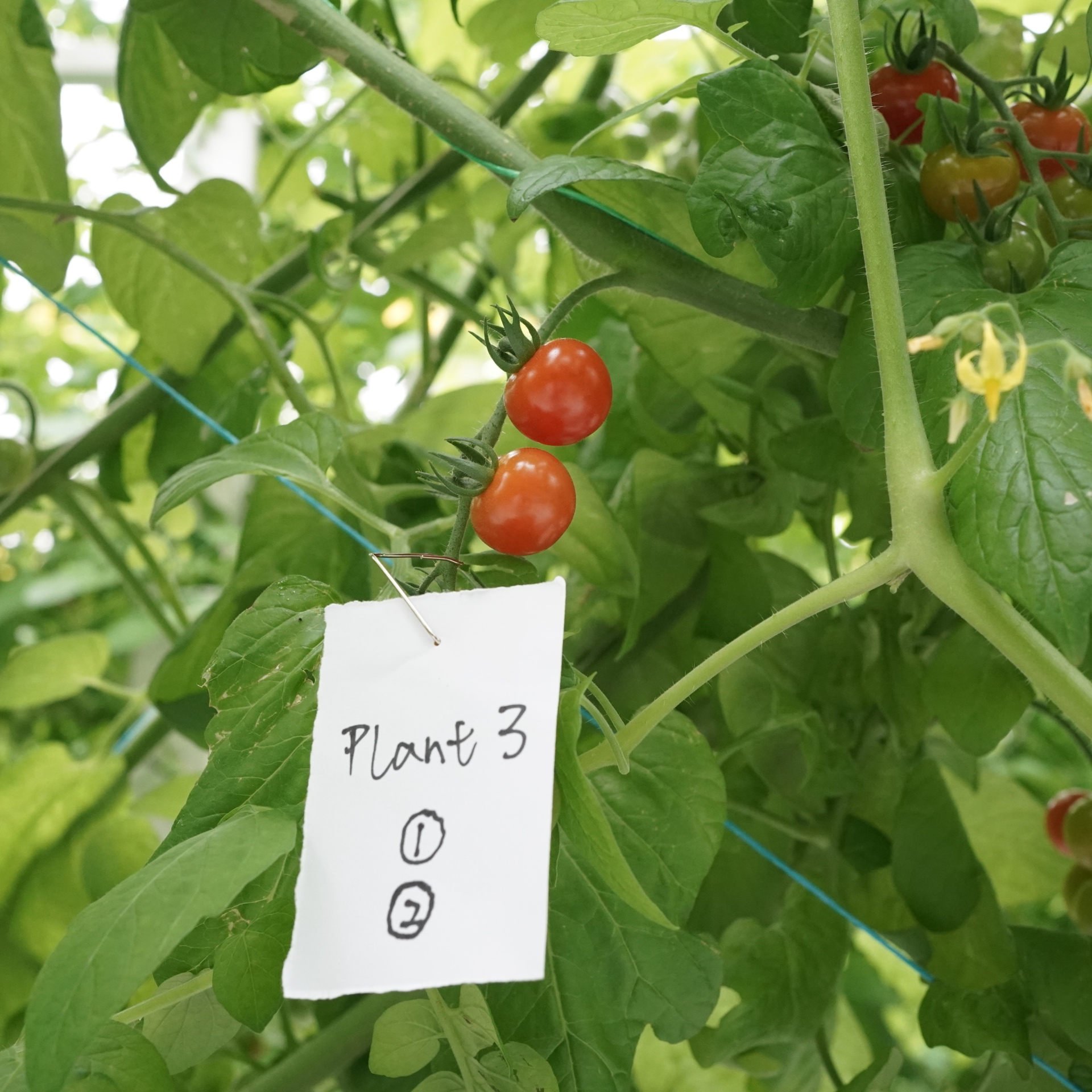}
        \caption{Input}
    \end{subfigure}
\hfill
    \begin{subfigure}{0.32\linewidth}
        \includegraphics[width=1.0\linewidth]{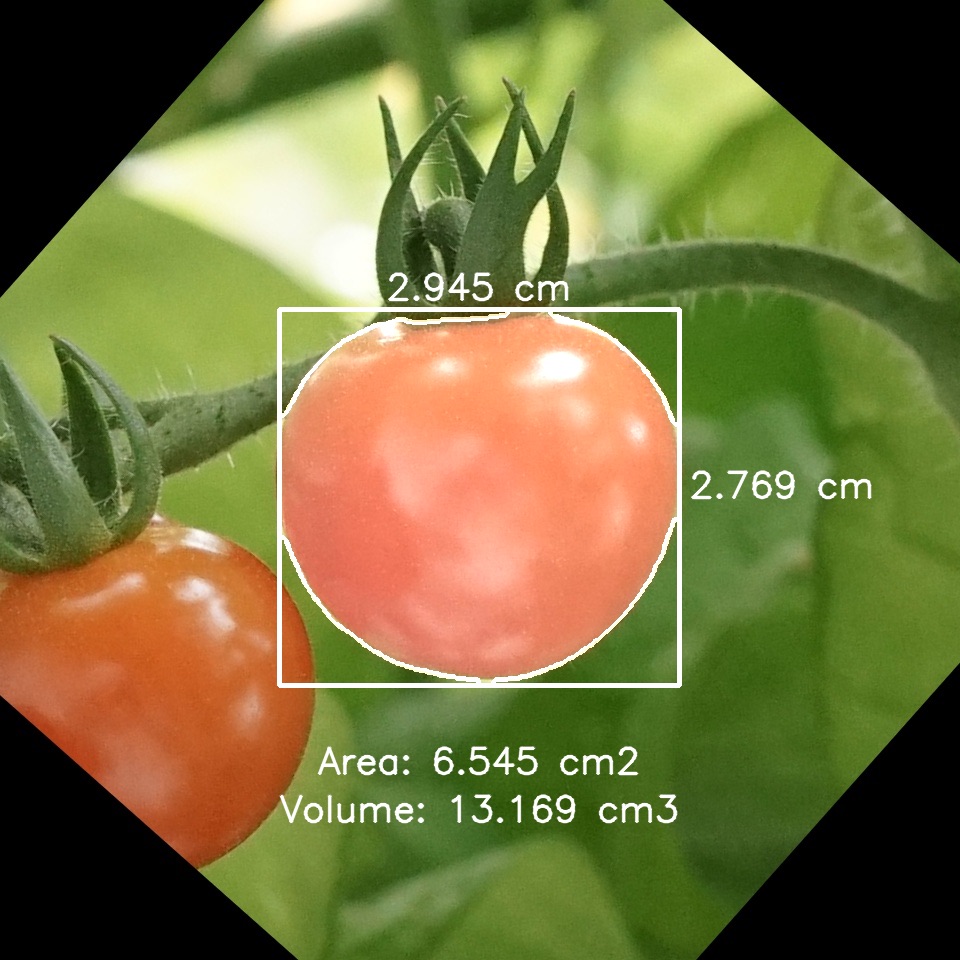}
        \caption{Output NO.1}
    \end{subfigure}
\hfill
    \begin{subfigure}{0.32\linewidth}
        \includegraphics[width=1.0\linewidth]{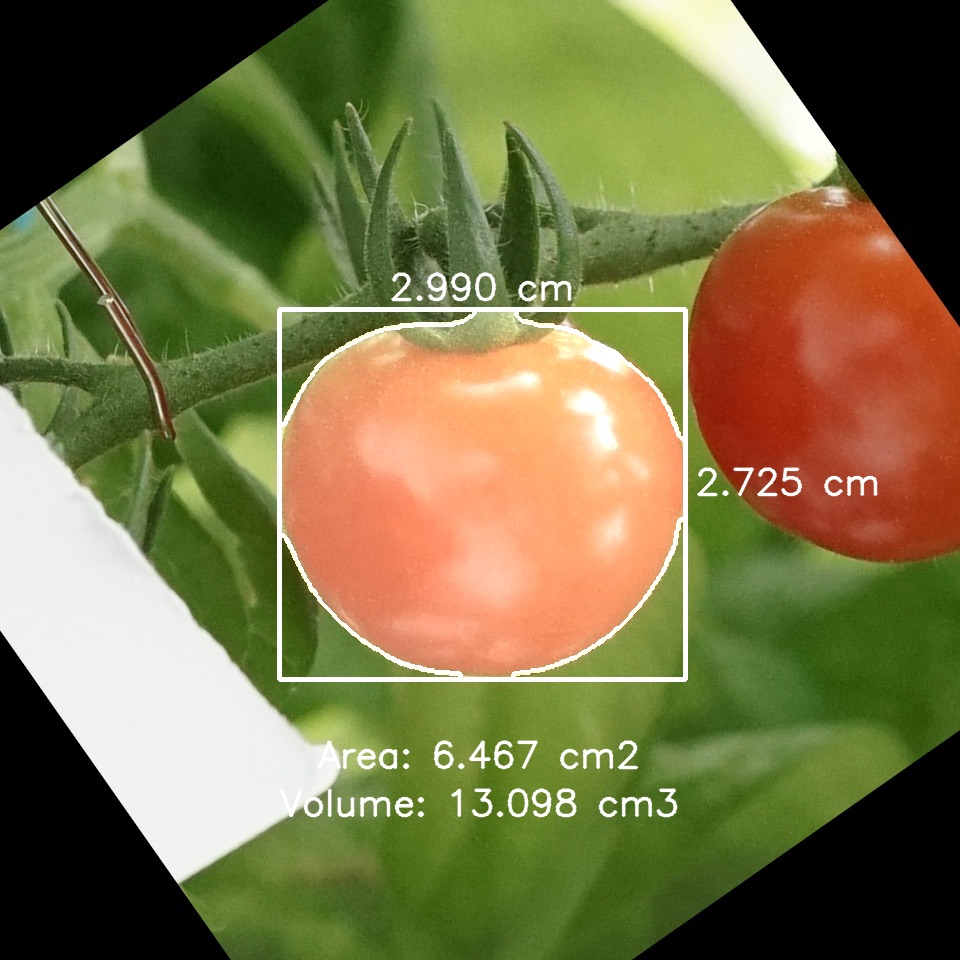}
        \caption{Output NO.2}
    \end{subfigure}
\caption{A simple demonstration of TomatoScanner: 
        (a) is the input RGB image. 
        (b) and (c) are the output phenotyping results - width, height, vertical area and volume - of two fruits, respectively. 
        (Zoom in for better observation)}
\label{fig:introduction_demo}
\end{figure}

\par Tomato is a delicious, nutritional and popular fruit, being loved by many people all over the world. 
Compared to traditional field planting, modern greenhouse planting achieves both better quality and higher yield, 
because environmental conditions can be precisely controlled to promote better crop growth \cite{greenhouse-meaning}. 
Phenotype refers to the morphological trait and growth state, such as leaf area, fruit volume, skin color, etc. 
Phenotypic measurement help researchers and farmers monitor in real time on how crops are growing, 
thereby make better decision on 
what task to implement, 
which plant to handle with, 
and when to do it \cite{phenotype-meaning}. 

\par Traditional phenotyping mainly relies on manual measurement via instruments such as vernier caliper and measuring glass, which is accurate but inefficient \cite{traditional-phenotyping}. 
More importantly, the greenhouse is full of sultry humid air with high concentrated CO$_2$, 
which seriously endangers the health and safety of workers \cite{heat-endanger-human}. 
In recent years, several studies have explored computer vision-based methods to replace manual phenotyping \cite{phenotype-survey}. 
However, the 2D-based need extra calibration \cite{2D-calibration}, or cause destruction to fruit \cite{2D-destructive}, or can only measure limited and meaningless traits \cite{2D-limit}. 
The 3D-based \cite{3D-reconstruction,3D-indoor,3D-leaf} need extra depth camera, which is expensive and unacceptable for most farmers. 
Thus, it is meaningful and urgent to develop a more advanced 2D-based method with excellent phenotyping accuracy, 
and deploy it on intelligent vehicle such as robotic dog. 

\par In this paper, we propose a non-contact tomato fruit phenotyping method, titled TomatoScanner, 
where RGB image is all you need for input, demonstrated in \cref{fig:introduction_demo}.
First, pixel feature is extracted by instance segmentation of our proposed EdgeYOLO with preprocessing of individual separation and pose correction. 
Second, depth feature is extracted by depth estimation of Depth Pro. 
Third, pixel and depth feature are fused to output phenotype results of four traits, 
including width (horizontal diameter), height (vertical diameter), vertical area and volume. 

\par We establish self-built Tomato Phenotype Dataset to test TomatoScanner, consisted of 25 image-phenotype pairs, where images and phenotypes are manually captured and measured, respectively. 
TomatoScanner achieves excellent phenotyping on width, height and vertical area, with median relative error of 5.63\%, 7.03\% and -0.64\%, respectively. 
In addition, we compare TomatoScanner with two state-of-the-art tomato phenotypic measurement methods. 
TomatoScanner greatly reduces input complexity, meanwhile, accomplishes the equally excellent phenotyping accuracy. 

\par EdgeYOLO is the core instance segmentation model in TomatoScanner. 
We propose and add three innovative modules - EdgeAttention, EdgeLoss and EdgeBoost - into EdgeYOLO, to enhance the segmentation accuracy on edge portion. 
In ablation experiment, if and only if all three modules are added, precision and mean Edge Error obtain the best improvement from 0.943 and 5.641\% to 0.986 and 2.963\%, respectively. 
Meanwhile, recall and mean Average Precision 50 only suffer tiny loss, proving the effectiveness of all three modules. 
In addition, we compare EdgeYOLO with five state-of-the-art instance segmentation models. 

\par The main contributions of this paper are as follows, where all codes and datasets are open-sourced: 
\begin{quote}
    1. TomatoScanner: a non-contact tomato phenotyping method, only need RGB image as input, which can be deployed on various downstream applications, such as robotic dog, monitor camera, interactive glasses and even smartphone \cite{guide-photo}. 
\end{quote}
\begin{quote}
    2. EdgeYOLO: an instance segmentation model, obtain improvement via addition of three innovative modules, which can be utilized for instance segmentation task in which accuracy of edge portion is more important. 
\end{quote}
\begin{quote}
    3. Tomato Phenotype Dataset: an self-built dataset, is consisted of 25 image-phenotype pairs, which can be utilized for test experiment in future studies of tomato phenotyping. 
\end{quote}

\par This paper is organized as follows: 
In \cref{sec:relatedworks}, related works of tomato phenotypic measurement and instance segmentation are introduced. 
In \cref{sec:methodology}, methodology of TomatoScanner and evaluation metrics are introduced. 
In \cref{sec:experiments}, dataset building, model training, test experiment, comparison experiment and ablation experiment are implemented. 
In \cref{sec:conclusion}, we conclude the advantages and shortages of TomatoScanner, then share the perspectives for future research. 

\section{Related Works}
\label{sec:relatedworks}

\subsection{Tomato Phenotypic Measurement}
\par Among all computer vision-based tomato phenotyping methods, there are two mainstreams: the 2D-based and the 3D-based. 
2D-based methods utilize RGB image as the only input, which is cheaper and more convenient for deployment. 
But they need extra calibration \cite{2D-calibration}, or cause destruction to fruit \cite{2D-destructive}, or can only measure limited and meaningless traits \cite{2D-limit}. 
3D-based methods introduce depth as extra input to implement 3D-reconstruction \cite{3D-reconstruction}, coordinate conversion \cite{3D-indoor} or 3D segmentation \cite{3D-leaf}, achieving excellent phenotyping accuracy. 
The economic burden of depth camera, however, is unacceptable for most farmers. 
Thus, it is meaningful and urgent to build a better 2D-based method with equal phenotyping accuracy. 

\subsection{Instance Segmentation}

\par Segmentation can be divided into two categories: semantic segmentation \cite{se-seg-Chen,se-seg-Liang,se-seg-Lin,se-seg-Ma,se-seg-Wu} and instance segmentation \cite{in-seg-Li,in-seg-Lian,in-seg-Wu,in-seg-Ying,in-seg-Zhang}. 
General instance segmentation models do not focus on specific portion of image. 
In our task, however, edge regions' accuracy are much more important than other portions' accuracy, 
because the following phenotyping process mainly rely on the edge region of segmentation mask. 
Thus, it is necessary and valuable to improve the edge region performance.

\section{Methodology}
\label{sec:methodology}

\par
\begin{figure*}
\centering
    \includegraphics[width=1.0\linewidth]{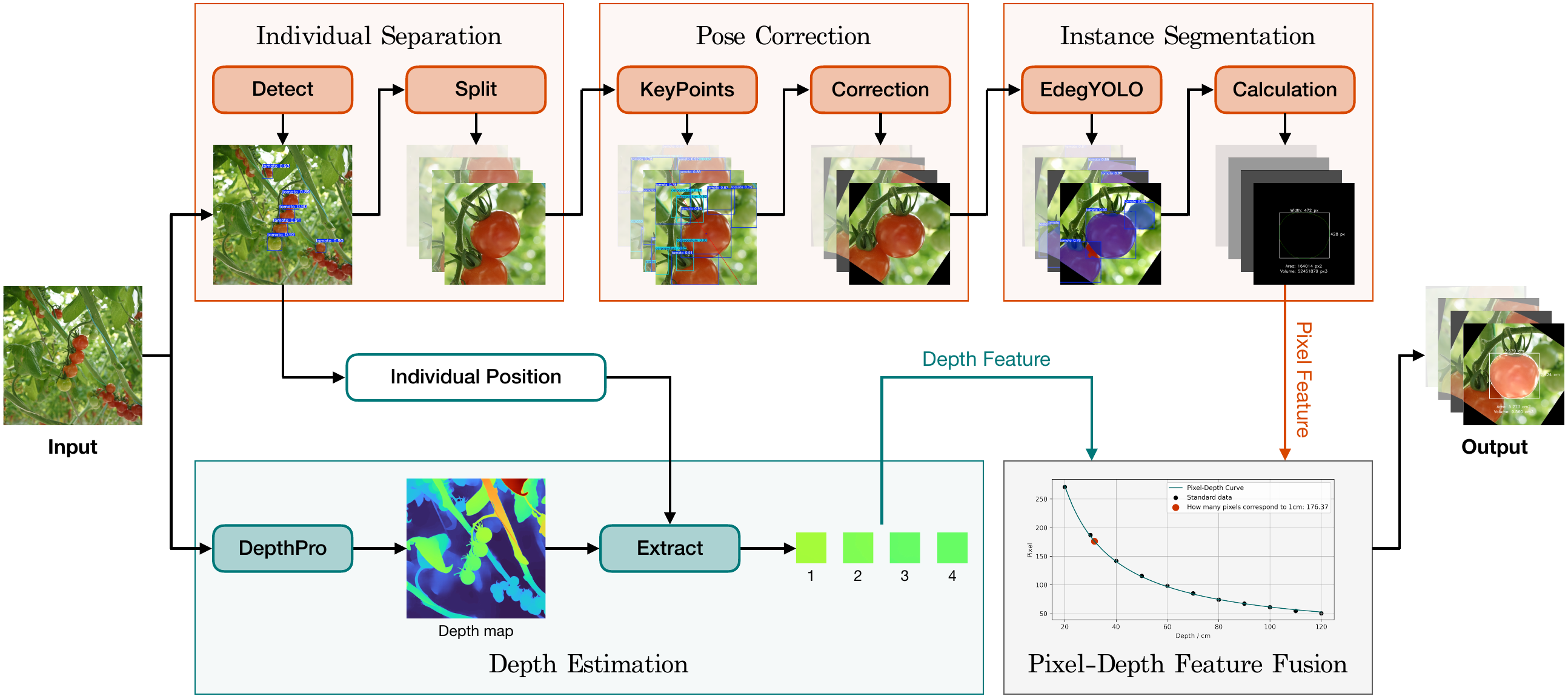}
    \caption{TomatoScanner architecture: Individual Separation module is illustrated in the top left. 
    Pose Correction module is illustrated in the top center. 
    Instance Segmentation module is illustrated in the top right. 
    Depth Estimation module is illustrated in the bottom left. 
    Pixel-Depth Feature Fusion module is illustrated in the bottom right. 
    (Zoom in for better observation)}
    \label{fig:TomatoScanner_architecture}
\end{figure*}

 We propose TomatoScanner to achieve state-of-the-art tomato phenotypic measurement, 
demonstrated in \cref{fig:TomatoScanner_architecture}. 
Pixel feature is extracted by EdgeYOLO introduced in Sec. 3.3, with preprocessing of individual separation introduced in Sec 3.1. and pose correction introduced in Sec. 3.2. 
Depth feature is extracted by Depth Pro introduced in Sec. 3.4. 
Pixel and depth feature are fused in Sec. 3.5 to output phenotypic results in reality. 

\subsection{Individual Separation}

\par In a bunch of tomatos, every fruit is in different size, angle and maturity. 
Therefore, every fruit should be measured independently. 
We utilize YOLO11s detector \cite{YOLO11} to detect all tomato fruits, then separate every fruit from the whole image to a regional image, 
demonstrated in the top left of \cref{fig:TomatoScanner_architecture}. 

\subsection{Pose Correcttion}
\par For a tomato fruit, correct pose refers to that carpopodium is directly above and body is directly below. 
If and only if being in correct pose, tomato fruit's pixel feature of phenotype can be validly extracted by EdgeYOLO. 
\par Pose correction is demonstrated in the top center of \cref{fig:TomatoScanner_architecture}. 
We utilize YOLO11s detector \cite{YOLO11} to detect two key points of tomato fruit: body and carpopodium. 
The pose vector $\mathbf{pose}$ is calculated by: 
\[
\mathbf{pose} = (x_c - x_b, y_c - y_b)
\]
where $(x_b, y_b)$ and $(x_c, y_c)$ are center positions of body and carpopodium, respectively. 
Fruit is corrected to vertical through rotation of image, in which the rotation angle $\theta$ is calculated by: 
\[
\theta = \cos^{-1} \left( \frac{\mathbf{pose} \cdot \mathbf{e}_y}{\|\mathbf{pose}\|} \right)
\]
where $\mathbf{e}_y$ is vertical unit vector. 
To keep image size unchanged, the blank background is filled with black color.

\subsection{Instance Segmentation}
\begin{figure*}
    \centering
    \includegraphics[width=1.0\linewidth]{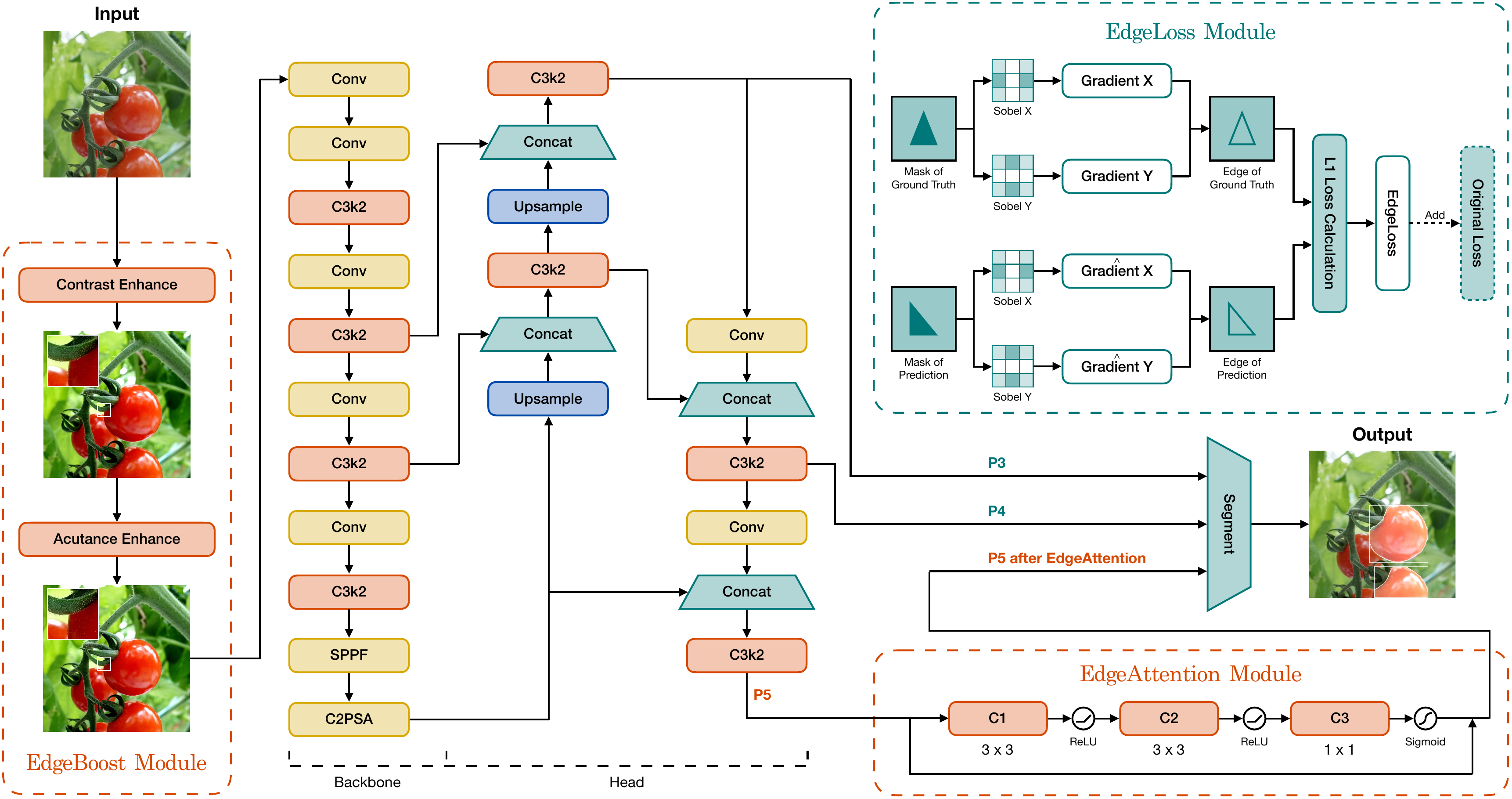}
    \caption{EdgeYOLO architecture: 
    Backbone and head are illustrated in the center; 
    EdgeAttention is illustrated in the bottom right; 
    EdgeLoss is illustrated in the top right; 
    EdgeBoost is illustrated in the bottom left.  
    (Zoom in for better observation)}
    \label{fig:EdgeYOLO_architecture}
\end{figure*}

We propose EdgeYOLO, an improved version of YOLO11m-seg, to achieve instance segmentation of tomato fruit, illustrated in \cref{fig:EdgeYOLO_architecture}. 
We implement three improvements - EdgeAttention, EdgeLoss and EdgeBoost - on the original model, 
to promote the segmentation performance on edge region of tomato fruit. 

\subsubsection{EdgeAttention}
We propose an innovative attention mechanism module, titled EdgeAttention, 
to pay more attention on edge region than global region, 
which is added after the lowest-resolution feature map $P5$.

\par EdgeAttention accepts $P5$ as input. 
To reduce channel dimension by half while capture structural details, $P5$ is fed to one $3 \times 3$ convolutional layer $C1$ with $ReLU$ activation to output initial feature. 
EdgeAttention uses another one $3 \times 3$ convolutional layer $C2$ with $ReLU$ activation to further enhance edge-specific feature, 
then uses one $1 \times 1$ convolutional layer $C3$ to produce single-channel attention map, which is normalized using $Sigmoid$ activation.
In the final, the original feature map is multiplied element-wise with the attention map, suppressing non-edge regions while amplifying edge-related responses.


\subsubsection{EdgeLoss}
We propose an innovative loss function, titled EdgeLoss, 
to additionally calculate the gap between predicted edge and ground truth edge, which is added into the original loss function. 
 
\par To calculate EdgeLoss value $\mathcal{L}_{edge}$, predicted mask $mask_{p}$ and ground truth mask $mask_{t}$ are taken as input. 
For every $mask_{t}$,
we use two $3 \times 3$ convolutional kernals - $ Sobel_x $ and $ Sobel_y $ - to compute the horizontal gradient $grad_x$ and the vertical gradient $grad_y$ of every pixel, respectively. 
$grad_x$ and $grad_y$ are merged to $grad$ by: 
\[
grad = \sqrt{{grad_x}^2 + {grad_y}^2}
\]
where $grad$ is the gradient magnitude of single pixel in $mask_t$. 
$mask_{p}$ is processed in the same way as above to output $\hat{grad}$. 
$\mathcal{L}_{edge}$ is calculated by L1 loss between $\hat{grad}$ and $grad$: 
\[
\mathcal{L}_{edge} = \frac{1}{N} \sum_{i=1}^{N} |\hat{grad}_i - {grad}_i|
\]
where $\hat{grad}_i$ or ${grad}_i$ is the gradient magnitude of pixel $i$ in $mask_p$ or $mask_t$, respectively. 
In the final, $\mathcal{L}_{edge}$ is added into the original loss value $\mathcal{L}_{original}$. 

\subsubsection{EdgeBoost}
We propose an innovative image preprocessing module, titled EdgeBoost, 
to boost edge regions standing out from the whole frame, 
which is added before the backbone. 

\par Image $I$ is processed through two steps: 
First, $I$ is enhanced on contrast with degree factor $\lambda _c$, 
where contrast refers to the difference in brightness between the light and dark areas. 
Second, $I$ is further enhanced on acutance with degree factor $\lambda _a$, 
where acutance measures the clarity of edges and fine details. 
In summary, EdgeBoost runs through: 
\[
I' = A[C(I, \lambda _c), \lambda _a]  
\]
where $I'$ is the processed image after EdgeBoost. 
Function $C$ and $A$ represent the step of contrast and acutance, respectively. 
After several experiments, $\lambda _c$ and $\lambda _a$ are set to 1.5 and 1.6 as default, respectively. 

\subsubsection{Phenotype Calculation}
\par We propose the phenotype calculation algorithm to calculate tomato fruit phenotype - width, height, vertical area and volume - based on the instance segmentation mask. 
\par Specifically, width is equal to the horizontal distance between the leftmost point and the rightmost point. 
Height is equal to the vertical distance between the topmost point and the bottommost point. 
Vertical area is equal to the area of the mask polygon. 
Volume is equal to the integral of the horizontal area with respect to \( y \), calculated by:
\[
V = \int_{y_{\text{min}}}^{y_{\text{max}}} \pi \left( \frac{D(y)}{2} \right)^2 \, dy
\]
where \( D(y) \) refers to the diameter of mask at the given y-coordinate, equal to the distance between the rightmost and the leftmost. 

\subsection{Depth Estimation}
\par We introduce Depth Pro, a zero-shot metric monocular depth estimation model proposed by Apple \cite{DepthPro}, to produce depth feature, demonstrated in \cref{fig:TomatoScanner_architecture}. 
Depth Pro utilize RGB image as the only input, then output the same-resolution depth map. 
According to every fruit's individual location from YOLO11s detector, Depth Pro extract the depth feature from the depth map for every individual fruit. 

\[
D_p = DepthMap(l_x, l_y)
\]
where $D_p$ represents predicted depth; 
$l_x$ and $l_y$ represent x-axis and y-axis of fruit's individual location, respectively. 

\subsection{Pixel-Depth Feature Fusion}
\par We propose Pixel-Depth Feature Fusion algorithm to fuse pixel feature from EdgeYOLO and depth feature from Depth Pro, outputting phenotypic results in reality, demonstrated in \cref{fig:TomatoScanner_architecture}.
Specifically, the 10cm calibration ruler is placed at various depth from 20cm to 120cm. 
Meanwhile, camera takes photos for every depth, and pixel quantity of calibration ruler is manully counted. 
Inverse proportional function is fitted based on all $pixel$ and $depth$: 
\[
pixel = \frac{k}{depth}
\]
where $k$ is inverse proportion coefficient. 
For the given $depth$, the calculated $pixel$ represents how many pixels correspond to $1cm$ in reality. 
Pixel and depth feature are fused by: 
\[
P_p = {P_x} \div {(\frac{k}{D_p})^j}
\]
where $j \in \{1, 2, 3\}$; \( P \) represents width ($j=1$), height ($j=1$), vertical area ($j=2$) or volume ($j=3$), respectively; 
\( P_p \), \( P_x \) and \( D_p \) represent predicted value of \( P \), pixel value of \( P \), and predicted value of depth, respectively. 

\subsection{Evaluation Metrics}
\subsubsection{for TomatoScanner}
We introduce Relative Error (RE) to evaluate phenotyping accuracy. 
For every phenotype denoted as \( P \), \( RE \) is calculated by: 
\[
P_e = \frac{P_p - P_t}{P_t} \times 100\%
\]
where \( P \) represents width, height, vertical area or volume, respectively; 
\( P_t \), \( P_p \) and \( P_e \) represent ground truth value, predicted value and \( RE \) of \( P \), respectively. 

\subsubsection{for EdgeYOLO}
We introduce 6 metrics to evaluate instance segmentation model:  
weights size, frame pre second (FPS), precision (P), recall (R), mean average precision 50 (mAP50) and mean edge error (mEE), 
where mEE is proposed by us to further focus on the segmentation accuracy on edge portion, 
because it is more important than else portion in our tomato phenotyping task. 

\begin{figure}[h]
    \centering
    \includegraphics[width=1.0\linewidth]{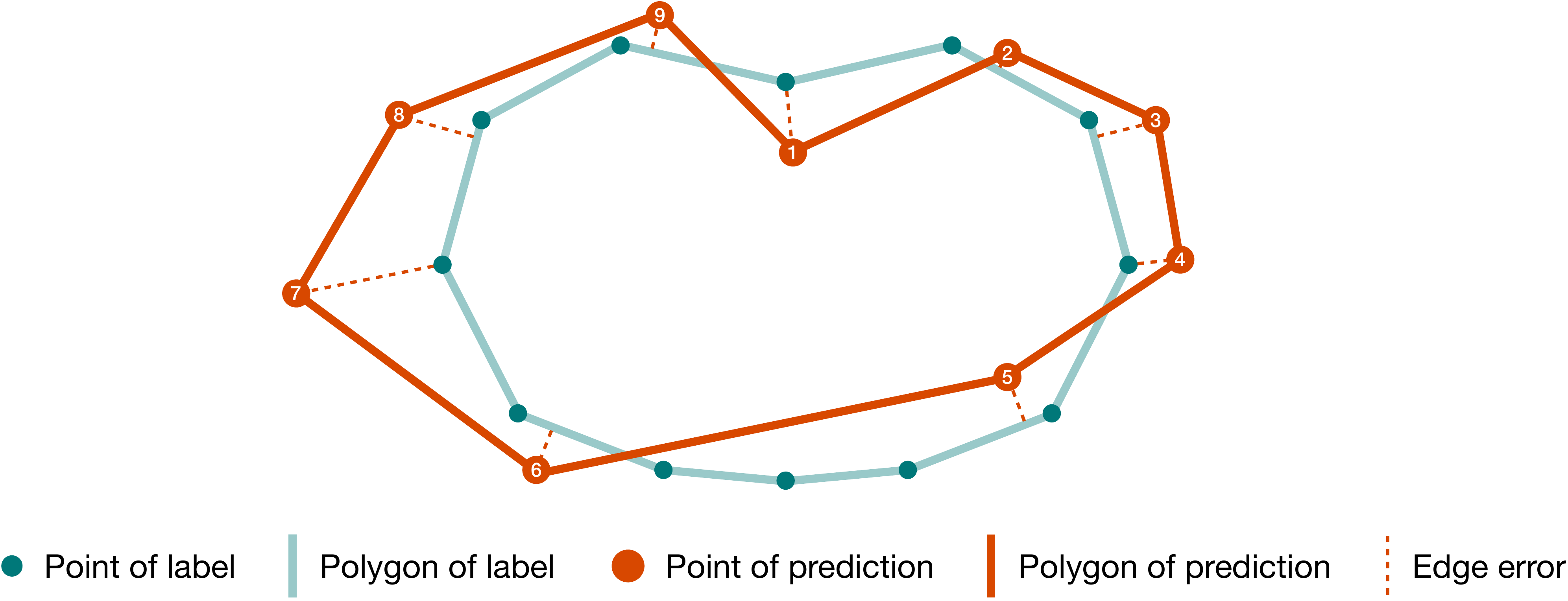}
    \caption{mEE calculation illustration 
    }
    \label{fig:mEE_illustration}
\end{figure}

The calculation of mEE is illustrated in \cref{fig:mEE_illustration}, where cyan mask is label and orange mask is prediction. 
Edge error of single predicted point refers to the minimum Euclidean distance from it to label polygon. 
mEE is calculated by averaging all edge error values in test dataset: 

\[
\text{mEE} = \frac{1}{M} \sum_{j=1}^{M} \left( \frac{1}{N_j} \sum_{i=1}^{N_j} d_{j,i} \right) \times 100\%
\]
where \( M \) is the number of images in the test dataset,  
\( N_j \) is the number of points sampled from the predicted mask of the \( j \)-th image,  
\( d_{j,i} \) represents the minimum Euclidean distance from the \( i \)-th sampled point in the \( j \)-th image to the ground truth mask boundary.

\section{Experiments}
\label{sec:experiments}

\subsection{Dataset}
\par We establish three self-built datasets for different tasks.
Tomato Phenotype Dataset is for evaluation of TomatoScanner. 
Tomato Detection Dataset is for training of YOLO11 detector. 
Tomato Segmentation Dataset is for training and evaluation of EdgeYOLO. 

\subsubsection{Tomato Phenotype Dataset}
\par
\begin{figure}[htbp]
  \centering
  \begin{subfigure}{0.48\linewidth}
   \includegraphics[width=1.0\linewidth]{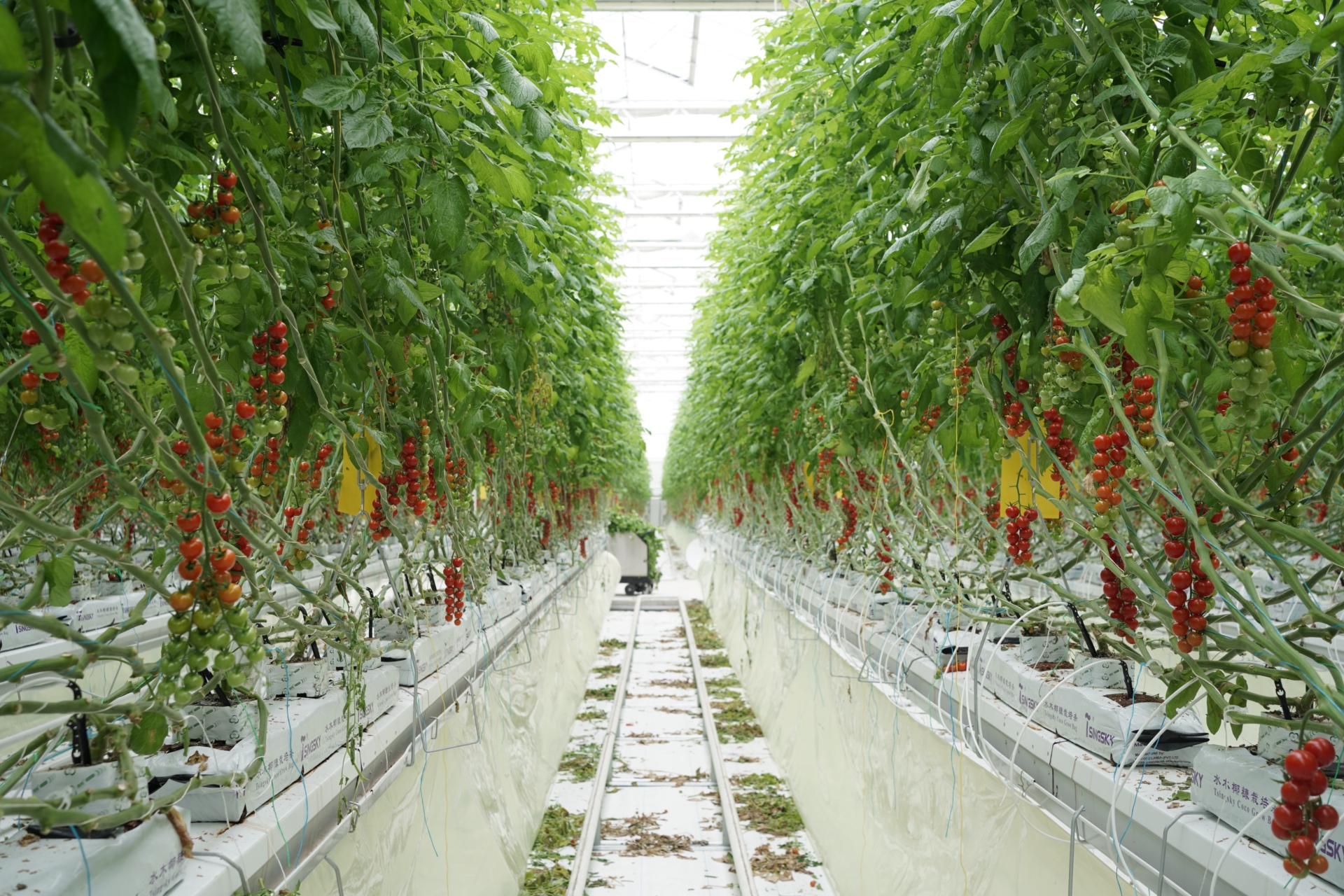}
    \caption{Tomato greenhouse}
  \end{subfigure}
  \hfill
  \begin{subfigure}{0.48\linewidth}
   \includegraphics[width=1.0\linewidth]{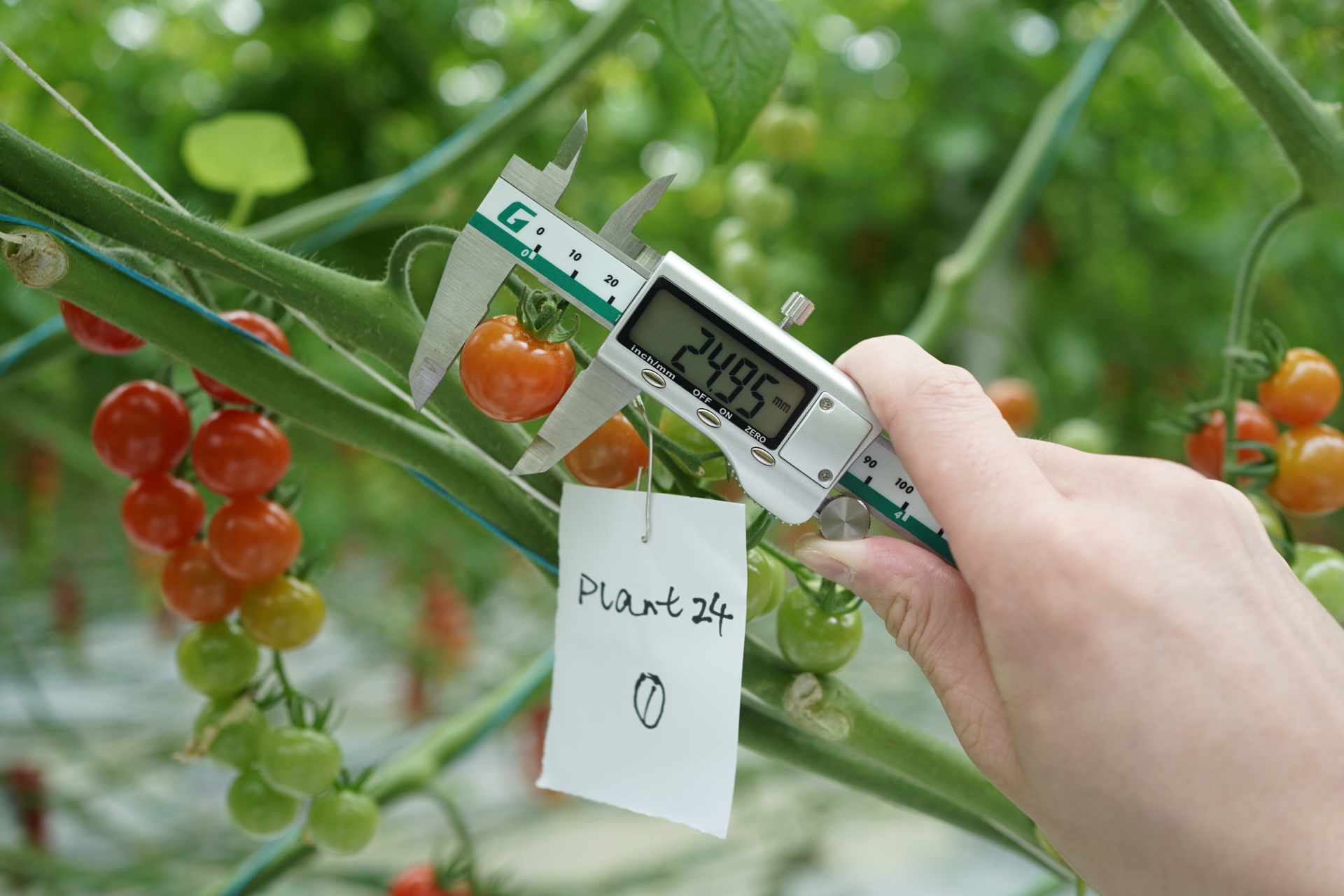}
   \caption{Manual measurement}
 \end{subfigure}
 \caption{Tomato Phenotype Dataset}
 \label{fig:tomato_phenotype_dataset}
\end{figure}

 A total of 25 image-phenotype pairs is manully collected in the greenhouse of Tsingsky Future Agricultural Science Center, China, 
demonstrated in \cref{fig:tomato_phenotype_dataset}. 
Image with $4000 \times 4000$ resolution is captured by Sony Alpha 7C camera with Sony FE 35mm F1.8 lens. 
Width and height are measured by vernier caliper. 
Area is measured by grid counting. 
Volume is measured by measuring cup drainage.

\subsubsection{Tomato Detection Dataset}
\par A total of 448 image-box pairs is collected. 
Web-crawler is introduced to download images efficiently. 
Make Sense app is utilized to annotate detection boxes manully. 
Data augmentation is applied to overcome the complex light condition in greenhouse \cite{Vaccine-YOLOv10}. 
Whole dataset is divided to train: val: test = 298: 100: 50.

\subsubsection{Tomato Segmentation Dataset}
\par A total of 381 image-mask pairs is collected. 
Web-crawler is introduced to download images efficiently. 
Labelme app is utilized to annotate segmentation masks manully. 
Whole dataset is divided to train: val: test = 261: 80: 40.

\begin{table*}[t]
  \centering
  \resizebox{\textwidth}{!}{
  \begin{tabular}{@{}cc|ccr|ccr|ccr|rrrl@{}}
    \toprule
    Plant & Fruit & $W_{t}$ & $W_{p}$ & $W_{e}$ & $H_{t}$ & $H_{p}$ & $H_{e}$ & $A_{t}$ & $A_{p}$ & $A_{e}$ & $V_{t}$ & $V_{p}$ & $V_{e}$ \\
    \midrule
    7  & 1 & 2.610 & 2.864 & +9.739  & 2.446 & 2.484 & +1.568  & 5.50 & 5.706 & +3.742  & 8 & 10.965  & +37.064 \\
    \midrule
    2  & 1 & 2.624 & 2.569 & -2.103  & 2.309 & 2.237 & -3.102  & 4.50 & 4.540 & +0.895  & 7 & 7.850   & +12.142 \\
       & 2 & 2.608 & 2.385 & -8.563  & 2.308 & 2.081 & -9.842  & 5.25 & 4.019 & -23.450 & 7 & 6.469   & -7.583 \\
    \midrule
    30 & 1 & 2.698 & 3.117 & +15.536 & 2.548 & 2.825 & +10.883 & 6.50 & 7.064 & +8.678  & 10 & 14.803 & +48.025 \\
       & 2 & 2.689 & 2.977 & +10.713 & 2.598 & 2.884 & +10.996 & 6.75 & 6.855 & +1.562  & 9 & 13.759  & +52.876 \\
       & 3 & 2.601 & 2.814 & +8.175  & 2.509 & 2.837 & +13.072 & 5.50 & 6.352 & +15.491 & 8 & 12.102  & +51.272 \\
       & 4 & 2.422 & 2.732 & +12.795 & 2.296 & 2.592 & +12.884 & 5.25 & 5.920 & +12.764 & 7 & 11.206  & +60.081 \\
       & 5 & 2.329 & 2.452 & +5.269  & 2.128 & 2.463 & +15.760 & 4.00 & 4.698 & +17.444 & 6 & 7.685   & +28.084 \\
       & 6 & 1.992 & 2.172 & +9.012  & 1.889 & 2.187 & +15.780 & 3.50 & 3.699 & +5.697  & 3 & 5.374   & +79.129 \\
    \bottomrule
  \end{tabular}}
  \caption{Test experiment results: 
  W, H, A and V refer to width, height, vertical area and volume, respectively. 
  Subscript t, p and e refer to ground truth value, predicted value and relative error, respectively. 
  Positive sign \& cyan background or negative sign \& orange background represent that predicted value is bigger or smaller than ground truth value, respectively. 
  The darker the backbone, the larger the error. 
  }
  \label{tab:test_table}
\end{table*}

\begin{figure*}[h]
  \centering
  \includegraphics[width=1.0\linewidth]{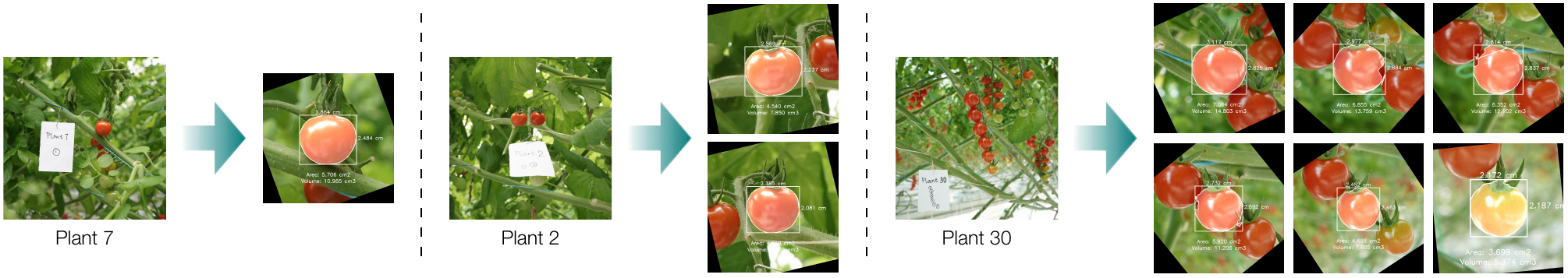}
  \caption{Test experiment demonstration: Left image is input. Right images are output. Predicted values of width, height, vertical area and volume are shown on the output images. 
  (Zoom in for better observation)}
  \label{fig:test_demonstration}
\end{figure*}

\subsection{Training}
\par EdgeYOLO is trained on Tomato Segmentation Dataset with 16 batch size.
YOLO11 detector is trained on Tomato Detection Dataset with 64 batch size.
Both trainings run on NVIDIA RTX 4090 (24 G video memory) with Python 3.11 for 300 epochs, 
illustrated in appendix \cref{fig:EdgeYOLO_training,fig:YOLO11_training}.

\subsection{Test Experiment}

\par We test TomatoScanner performance on Tomato Phenotype Dataset. 
No.7, No.2, No.30 are selected as the representatives, demonstrated in \cref{tab:test_table} and \cref{fig:test_demonstration}. 
The complete results are recorded in appendix \cref{tab:complete_test_table}. 
An RGB image is all we need for input, then phenotypic measurement results of all fruits are successfully outputted, including width, height, vertical area and volume. 
Instance segmentation masks and phenotypic measurement results are drawn on the output images for better visualization. 
For example, plant NO.2 has two fruits, in which the first fruit is predicted to 2.569 cm width and 7.850 cm$^3$ volume. 
The ground truths are 2.624 cm and 7 cm$^3$, respectively. 
So the relative errors are -2.103 \% and +12.142 \%, respectively. 

\par
\begin{figure}[b]
  \centering
  \includegraphics[width=1.0\linewidth]{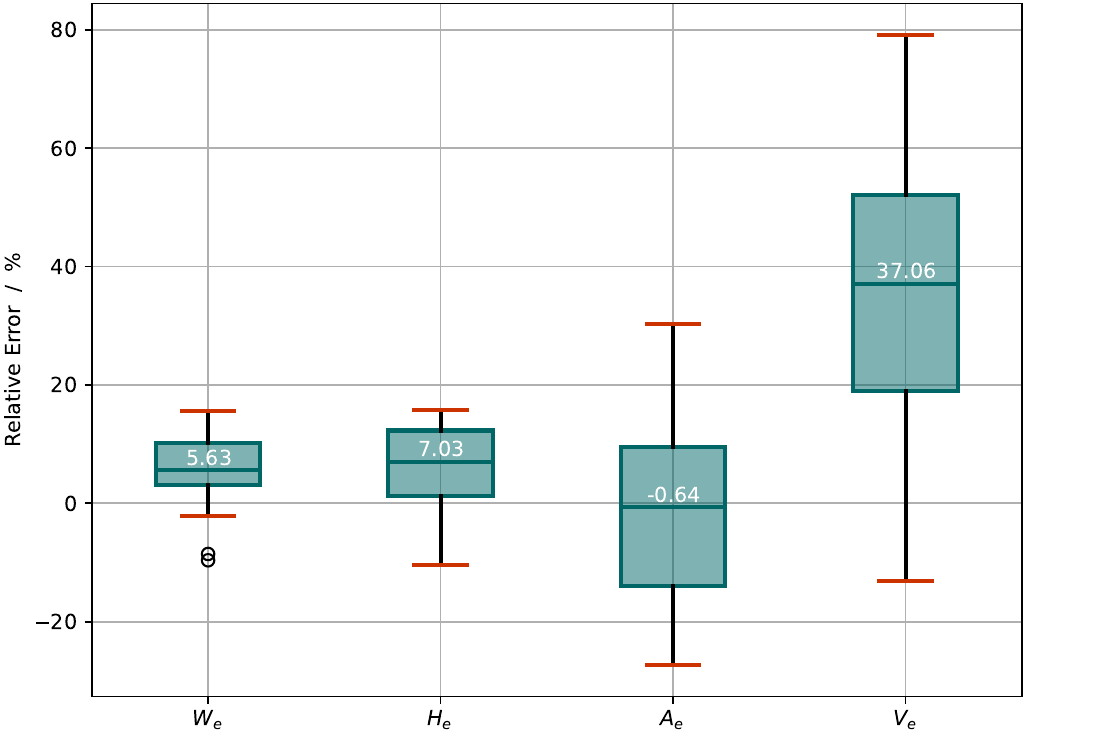}
  \caption{Test experiment statistic box plot: 
          median values are marked at corresponding positions. 
  }
  \label{fig:test_box_plot}
\end{figure}

 The statistic box plot of all relative errors is illustrated in \cref{fig:test_box_plot}.
For width and height, the madians are only 5.63 \% and 7.03 \%, respectively, 
meanwhile, the body boxes are narrow, which represents the outstanding accuracy and stability. 
For vertical area, the median is almost perfect with only -0.64 \%, 
however, the absolute maximum is over 20\% and the body box is long, which represents the excellent accuracy with a little unstability. 
For volume, the median is more than 30\%, which represents the inferior phenotypic measurement performance. 

\par To further explain the measurement accuracy distinction between different phenotypic classification, we combine the calculation principle of each. 
Width, height and vertical area are directly extracted from the segmentation mask. 
Volume, however, is calculated by the integral of horizontal crossing area to vertical axis, where multi parameters may introdece errors. 
In the final, the cumulation of errors from level to level leads to the ultimate great error. 

\par In summary, TomatoScanner achieves excellent performance on width, height and vertical area in test experiment, 
which proves the effectiveness of our methodology. 
Volume measurement is currently inferior, deserving our further improvement. 

\subsection{Comparison Experiment}

\subsubsection{Tomato Phenotyping Methods}

\par We compare TomatoScanner with two state-of-the-art tomato phenotypic measurement methods, 
where Z. Gu et al.'s (ZG) \cite{2D-calibration} and T. Zhu et al.'s (TZ) \cite{3D-reconstruction} are selected as 2D-based and 3D-based methods, respectively. 
The comparison result is shown in \cref{tab:tomato_phenotyping_method_comparison_table}, 
including basic information and four aspects. 

\par Input complexity: Compared to ZG and Ours, TZ need extract depth camera to capture depth image, 
which brings additional economy cost. 

\par Phenotyping accuracy: For two common phenotype, width and height, 
although the evaluation matrics are different, all three methods achieves excellent scores. 

\begin{table}[b]
  \centering
  \resizebox{0.47\textwidth}{!}{ 
  \begin{tabular}{@{}c|c|c|cl@{}}
    \toprule
    Method       & Z. Gu et al.'s& T. Zhu et al.'s & Ours \\
    \midrule
    dimension    & 2D            & 3D              & 2D \\
    target       & fruit         & canopy          & fruit \\
    \midrule
    
    input image  & RGB           & RGB \& depth    & RGB \\
    device       & RGB camera    & RGB-D camera    & RGB camera \\
    complexity   & low           & high            & low \\
    \midrule

    width error  & 0.68 RMSE     & 1.92 cm(a)      & 0.08 RE(m) \\
    height error & 0.83 RMSE     & 1.38 cm(a)      & 0.09 RE(m) \\
    accuracy     & excellent     & excellent       & excellent \\
    \midrule

    condition    & calibration   & specific angle  & any \\
    process      & auto          & auto            & auto \\
    automation   & semi          & semi            & full \\
    \midrule
    
    core tech    & detection     & 3D-             & instance seg. \\
                 &               & reconstruction  & depth esti. \\
    computation  & low           & high            & middle \\
    \bottomrule
  \end{tabular}}
  \caption{Tomato phenotyping method comparison results: (a), (m), seg. and esti. refer to average, median, segmentation and estimation, respectively.}
  \label{tab:tomato_phenotyping_method_comparison_table}
\end{table}

\begin{table}[t]
   \centering
   \resizebox{0.47\textwidth}{!}{ 
   \begin{tabular}{@{}crrcc@{}}
     \toprule
     Model       & Weights & FPS   & mAP50 & mEE \\
     \midrule
     Mask R-CNN  & 334.8 M & 24.20 & 0.954 & 4.440 \% \\
     YOLACT      & 189.8 M & 15.83 & 0.923 & 6.149 \% \\
     RTMDet      &   1.6 G &  2.53 & 0.852 & 3.911 \% \\
     SAM2        & 154.4 M & 20.91 &     / & 1.377 \% \\
     YOLO11n-seg &   6.0 M & 66.37 & 0.969 & 5.436 \% \\
     YOLO11x-seg & 124.8 M & 55.56 & 0.962 & 6.271 \% \\
     Ours        &  48.7 M & 76.34 & 0.955 & 2.963 \% \\
     \bottomrule
   \end{tabular}}
   \caption{Instance segmentation model comparison results}
   \label{tab:instance_segmentation_model_comparison_table}
\end{table}

\begin{figure}[h]
  \centering
  \captionsetup[subfigure]{labelformat=empty}
  \begin{subfigure}{0.24\linewidth}
   \includegraphics[width=1.0\linewidth]{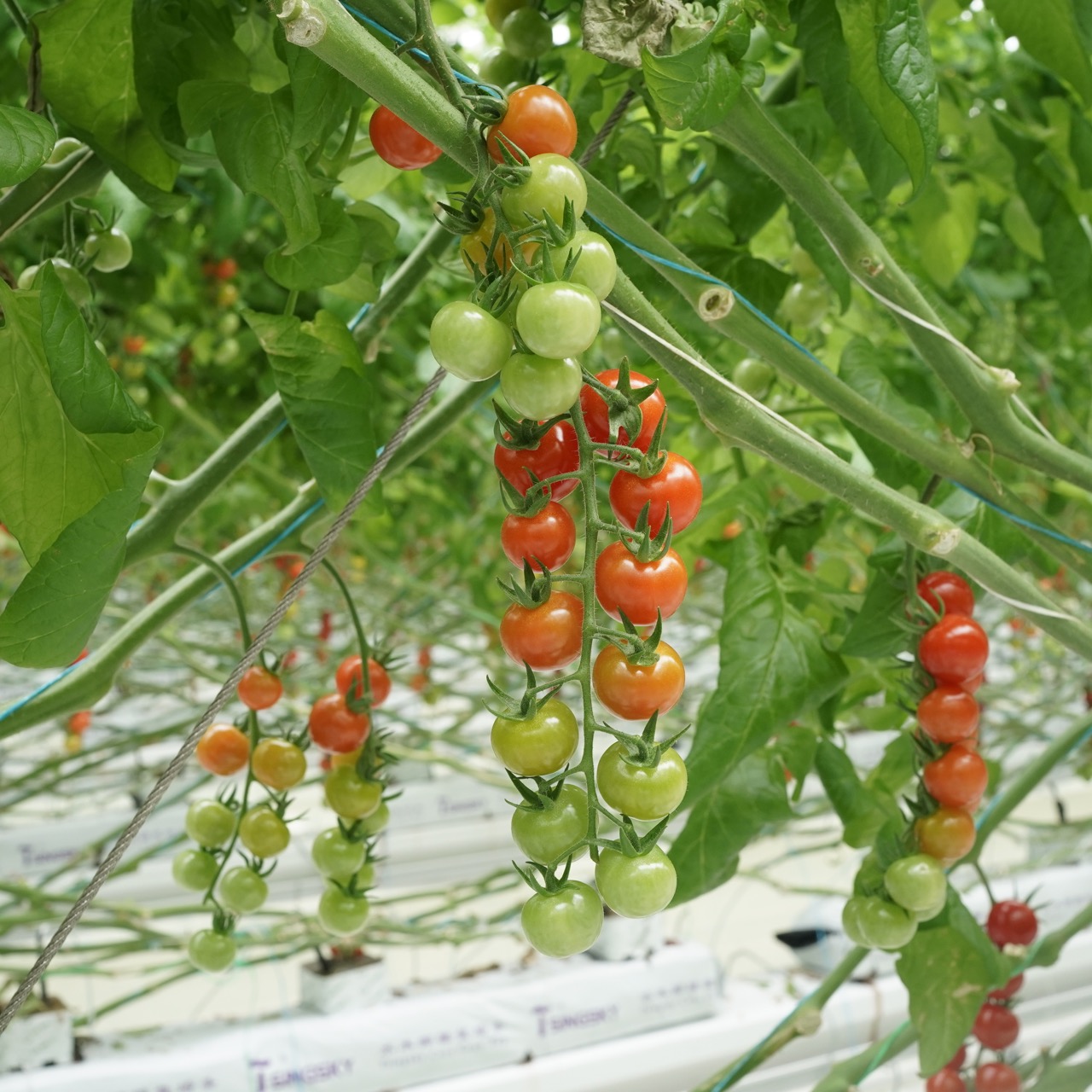}
    \caption{Input NO.1}
  \end{subfigure}
  \hfill
  \begin{subfigure}{0.24\linewidth}
   \includegraphics[width=1.0\linewidth]{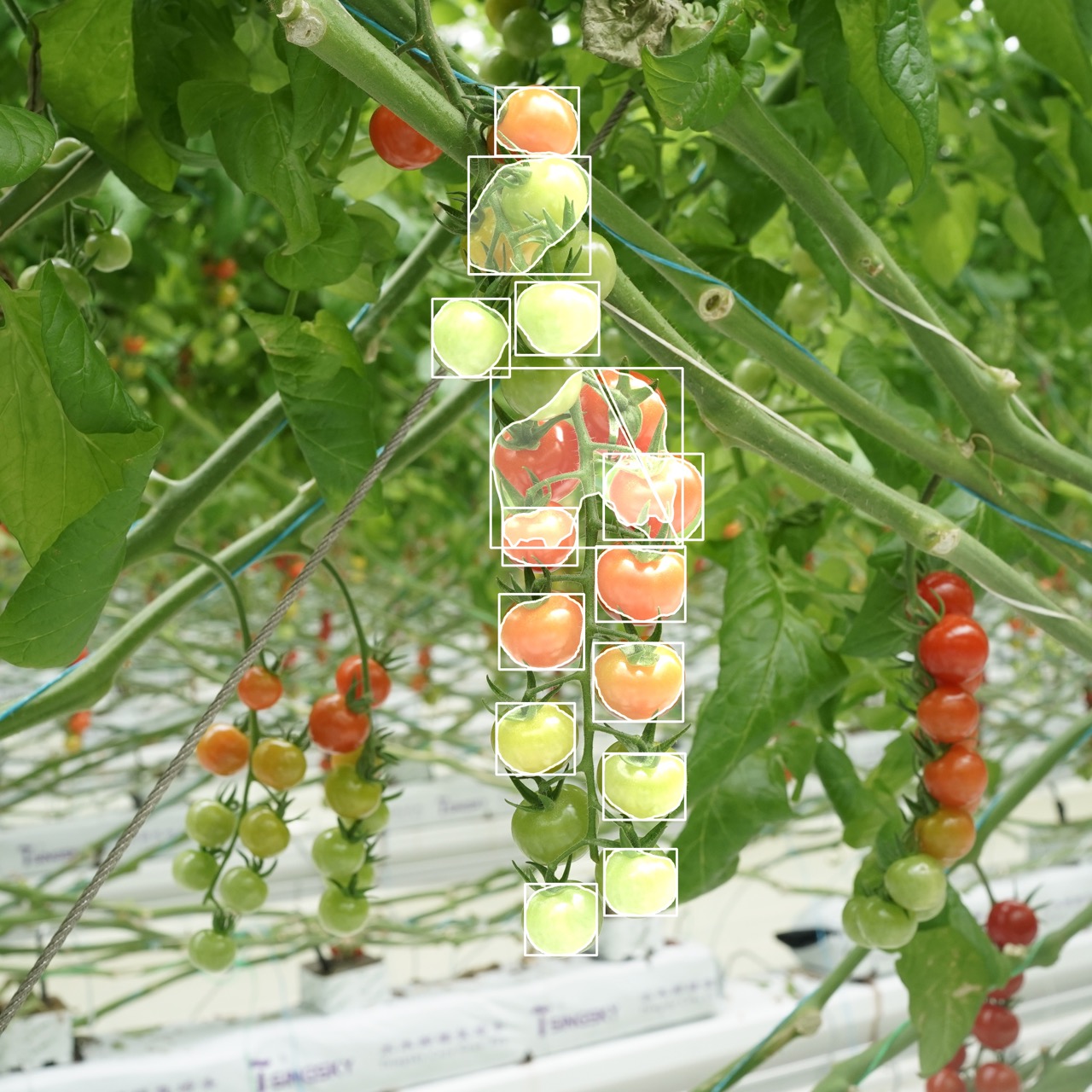}
   \caption{Mask R-CNN}
 \end{subfigure}
 \hfill
 \begin{subfigure}{0.24\linewidth}
   \includegraphics[width=1.0\linewidth]{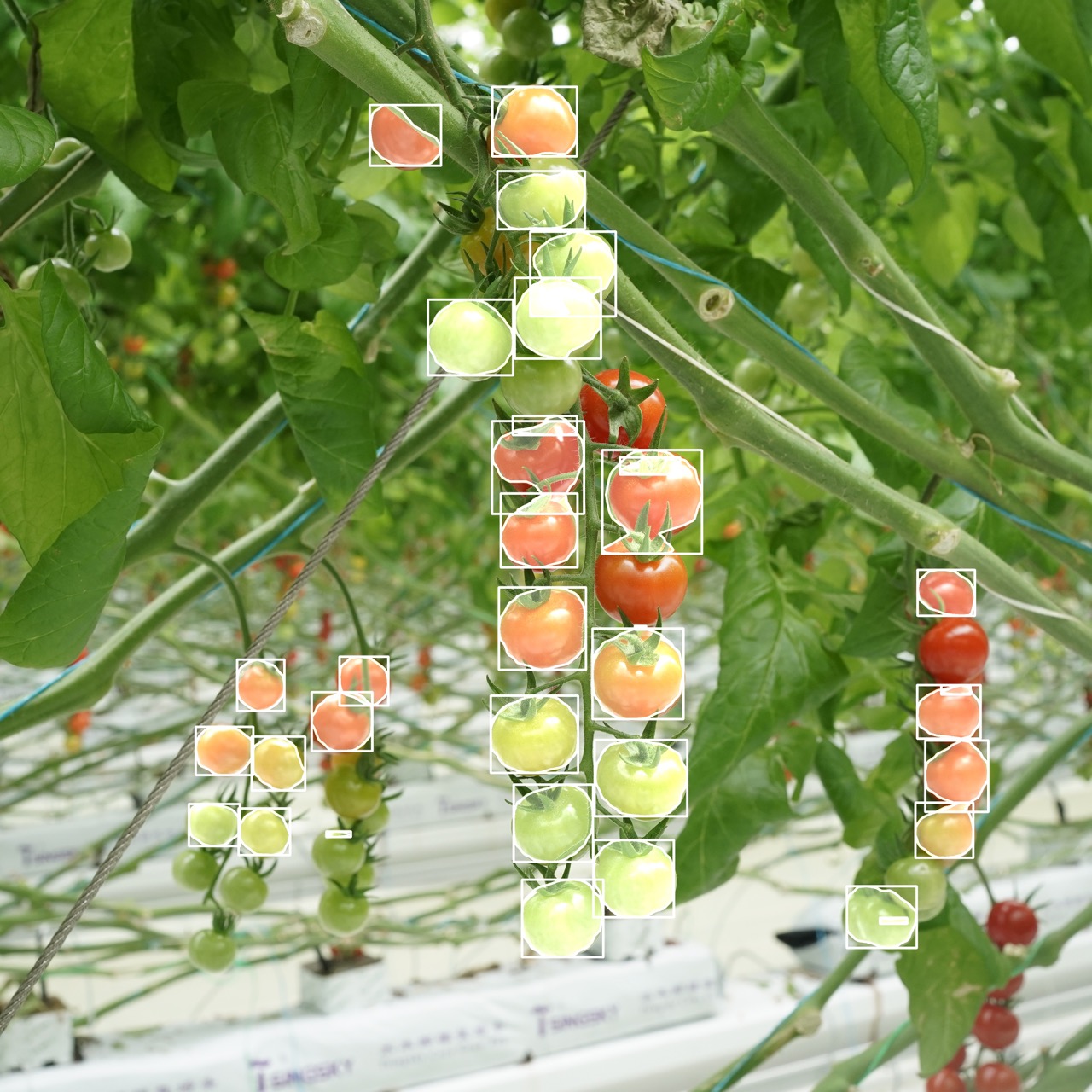}
   \caption{YOLACT}
 \end{subfigure}
 \hfill
  \begin{subfigure}{0.24\linewidth}
   \includegraphics[width=1.0\linewidth]{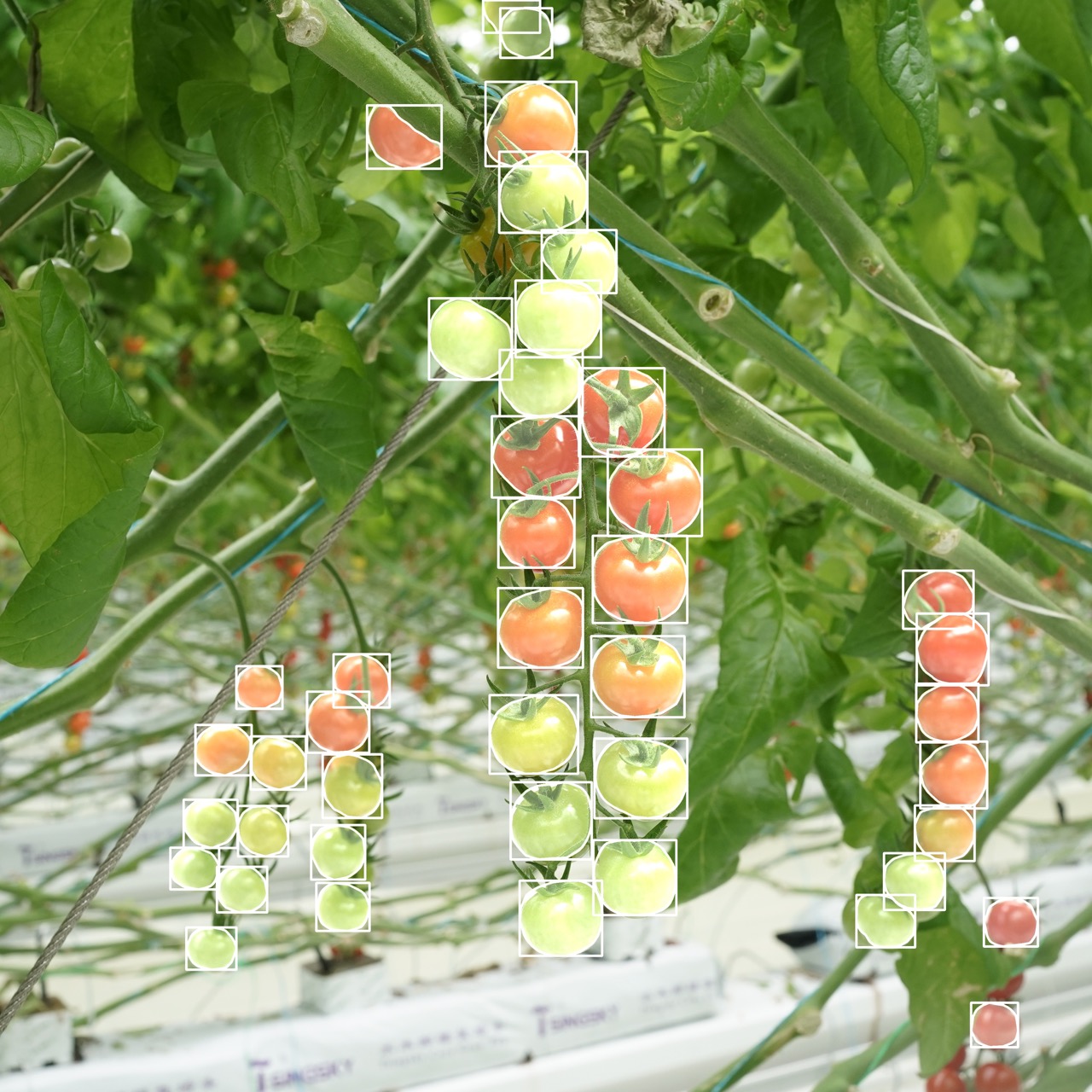}
    \caption{RTMDet}
  \end{subfigure}

  \begin{subfigure}{0.24\linewidth}
   \includegraphics[width=1.0\linewidth]{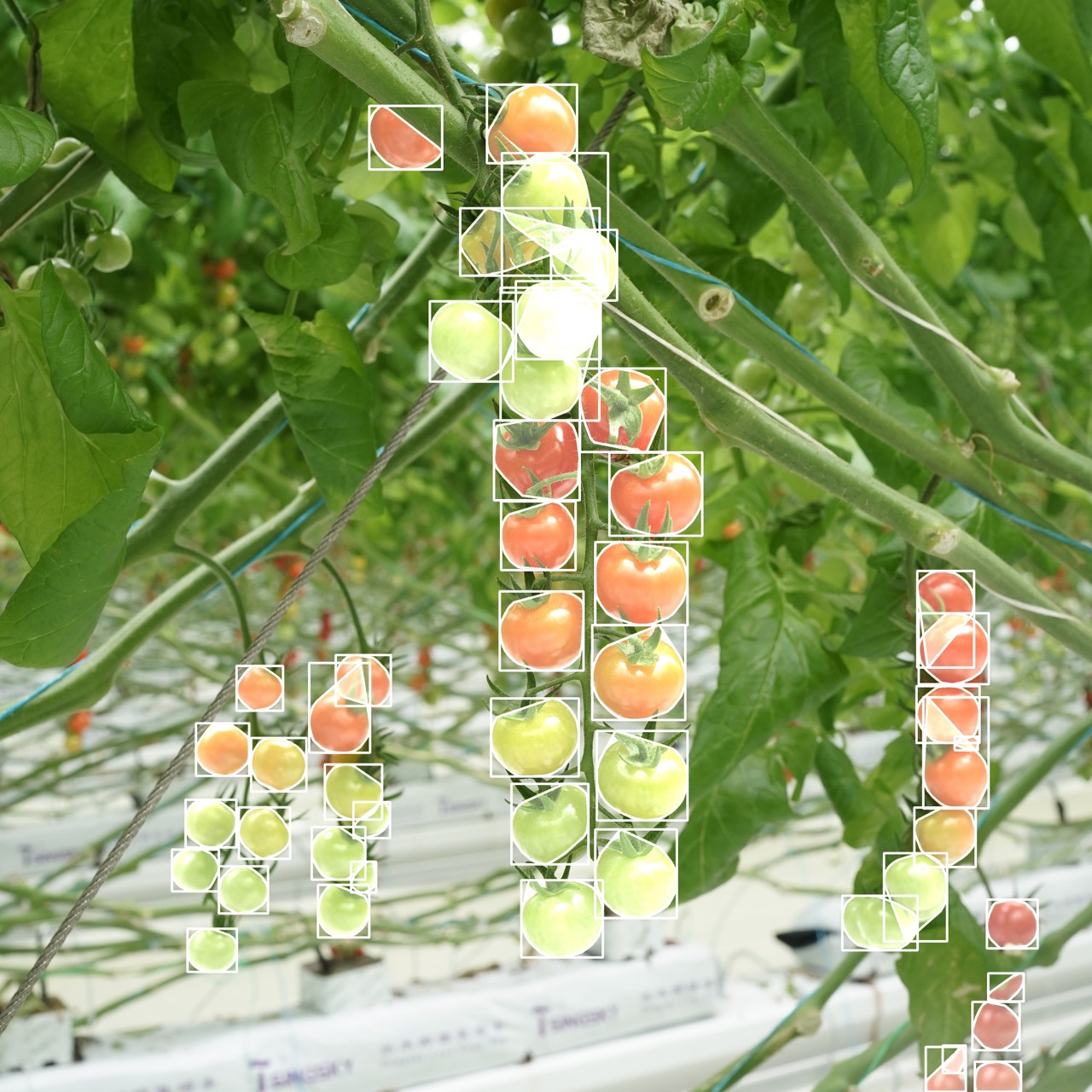}
    \caption{SAM2}
  \end{subfigure}
  \hfill
  \begin{subfigure}{0.24\linewidth}
   \includegraphics[width=1.0\linewidth]{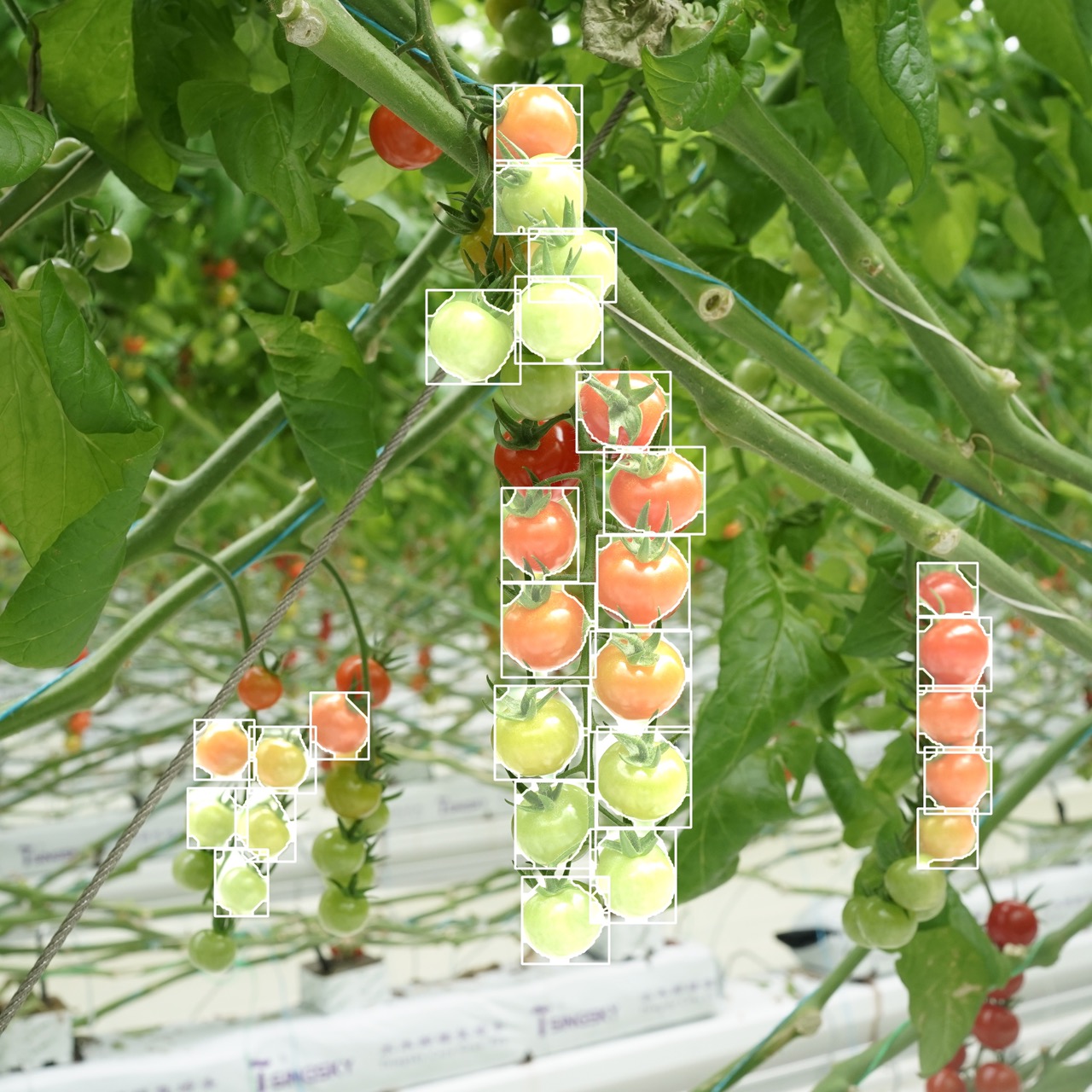}
    \caption{YOLO11n-seg}
  \end{subfigure}
  \hfill
  \begin{subfigure}{0.24\linewidth}
   \includegraphics[width=1.0\linewidth]{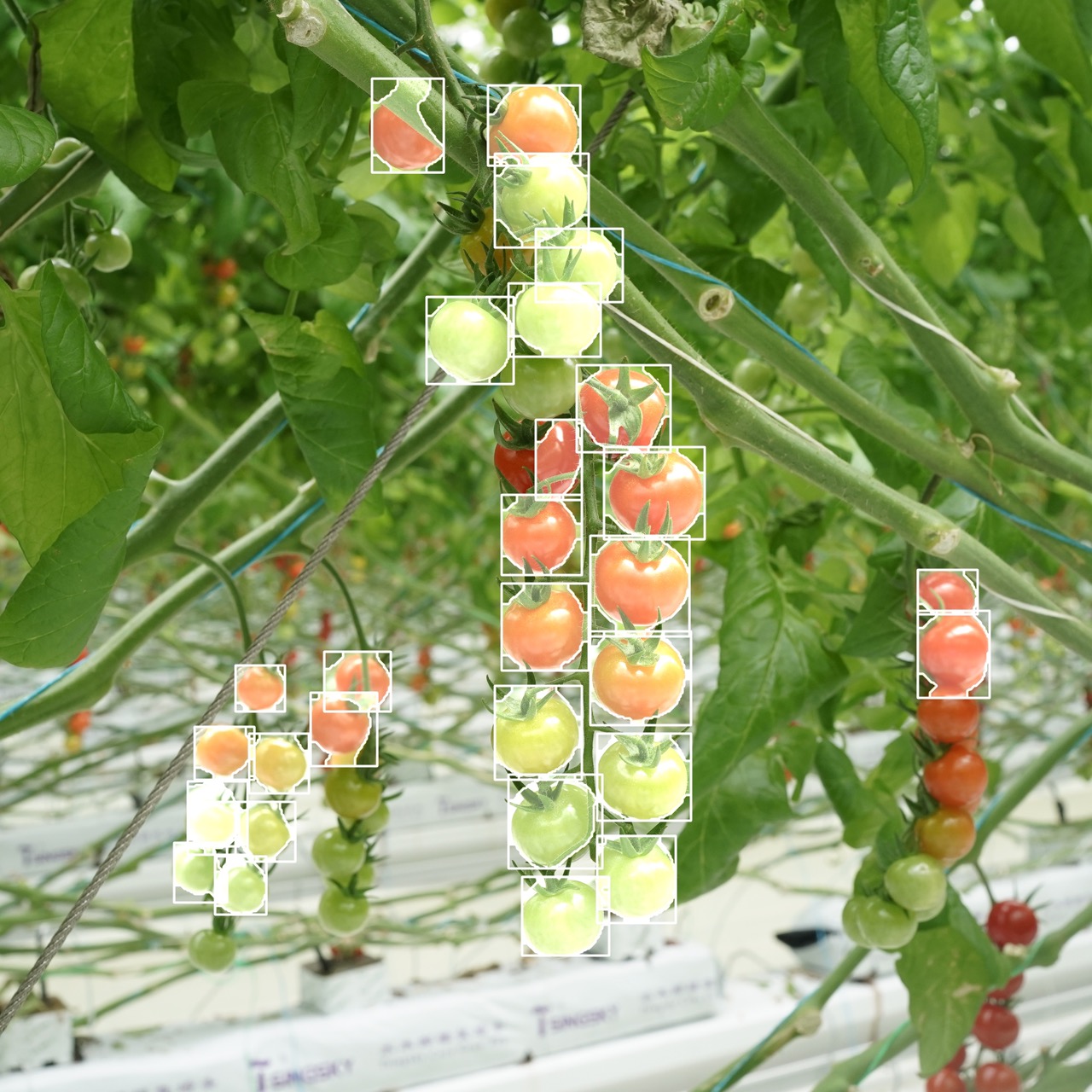}
    \caption{YOLO11x-seg}
  \end{subfigure}
  \hfill
  \begin{subfigure}{0.24\linewidth}
   \includegraphics[width=1.0\linewidth]{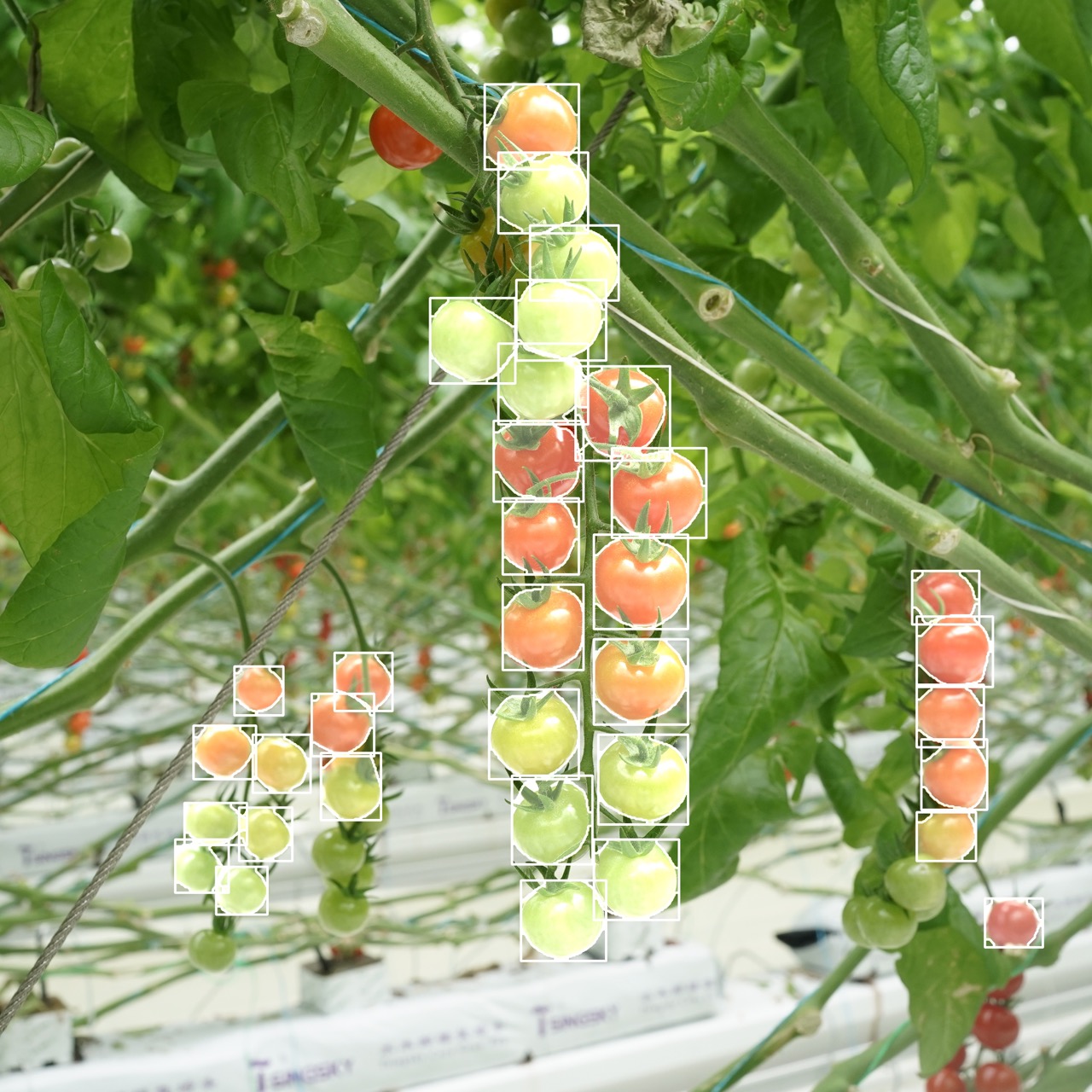}
    \caption{Ours}
  \end{subfigure}

  \begin{subfigure}{0.24\linewidth}
    \includegraphics[width=1.0\linewidth]{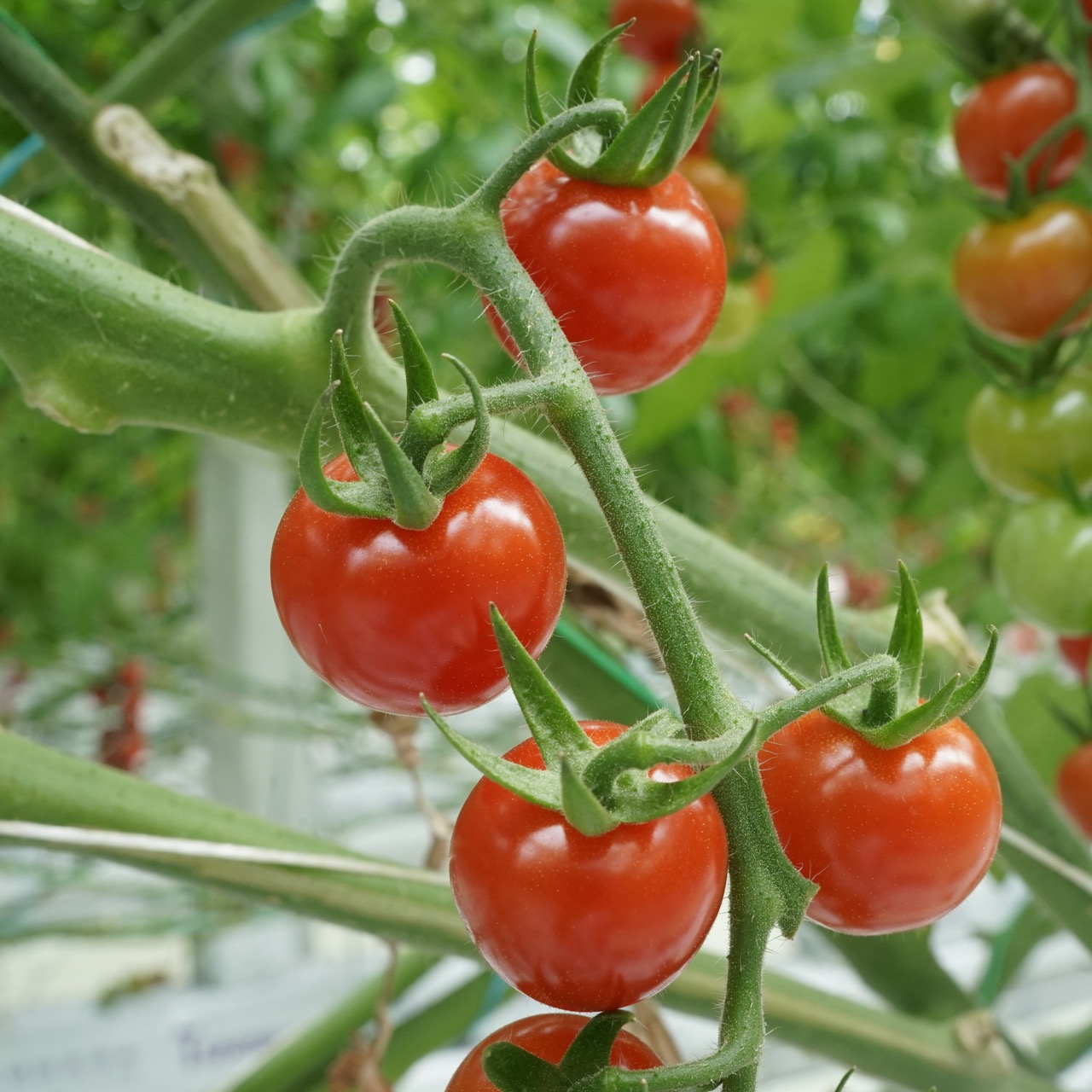}
     \caption{Input NO.2}
   \end{subfigure}
   \hfill
   \begin{subfigure}{0.24\linewidth}
    \includegraphics[width=1.0\linewidth]{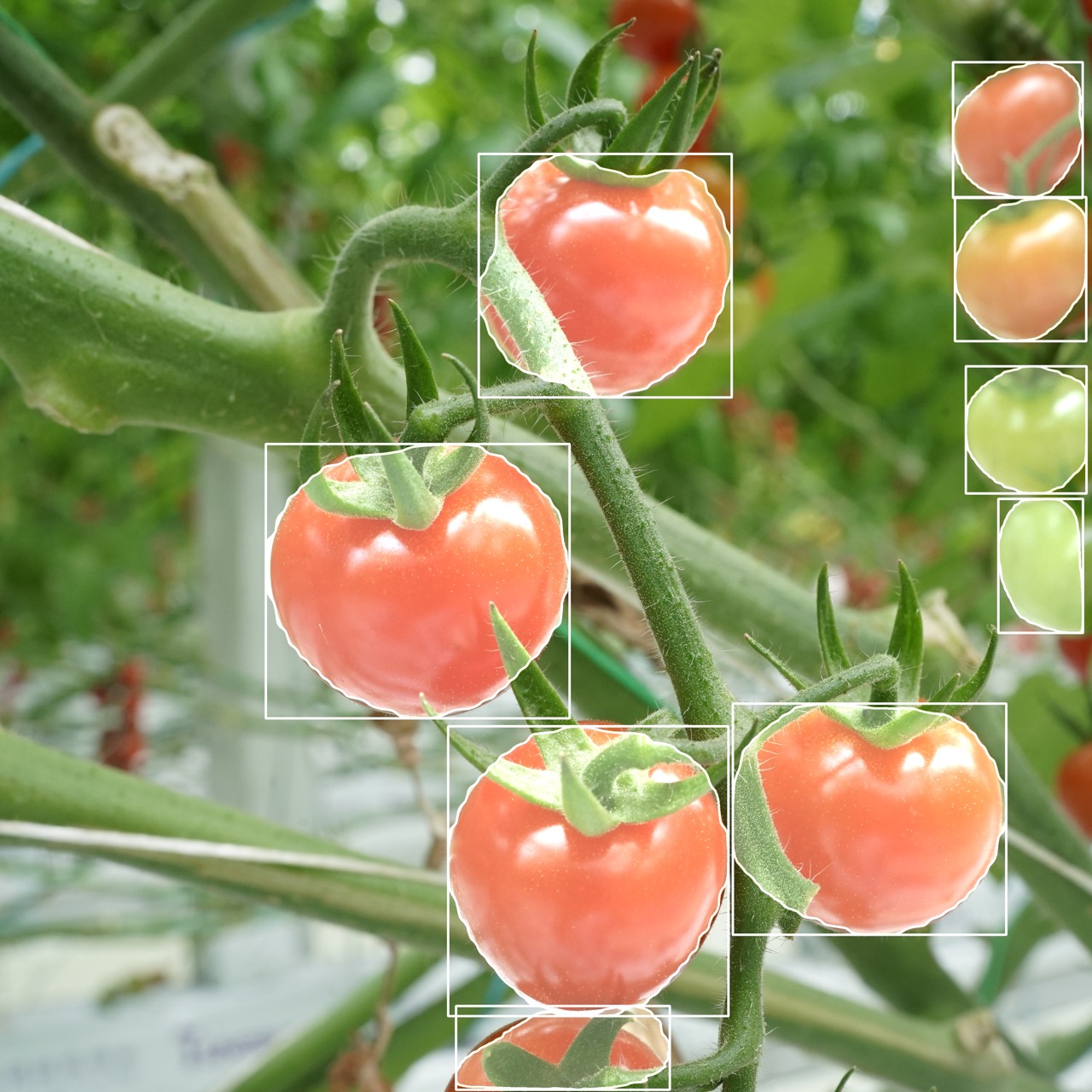}
     \caption{Mask R-CNN}
   \end{subfigure}
   \hfill
   \begin{subfigure}{0.24\linewidth}
    \includegraphics[width=1.0\linewidth]{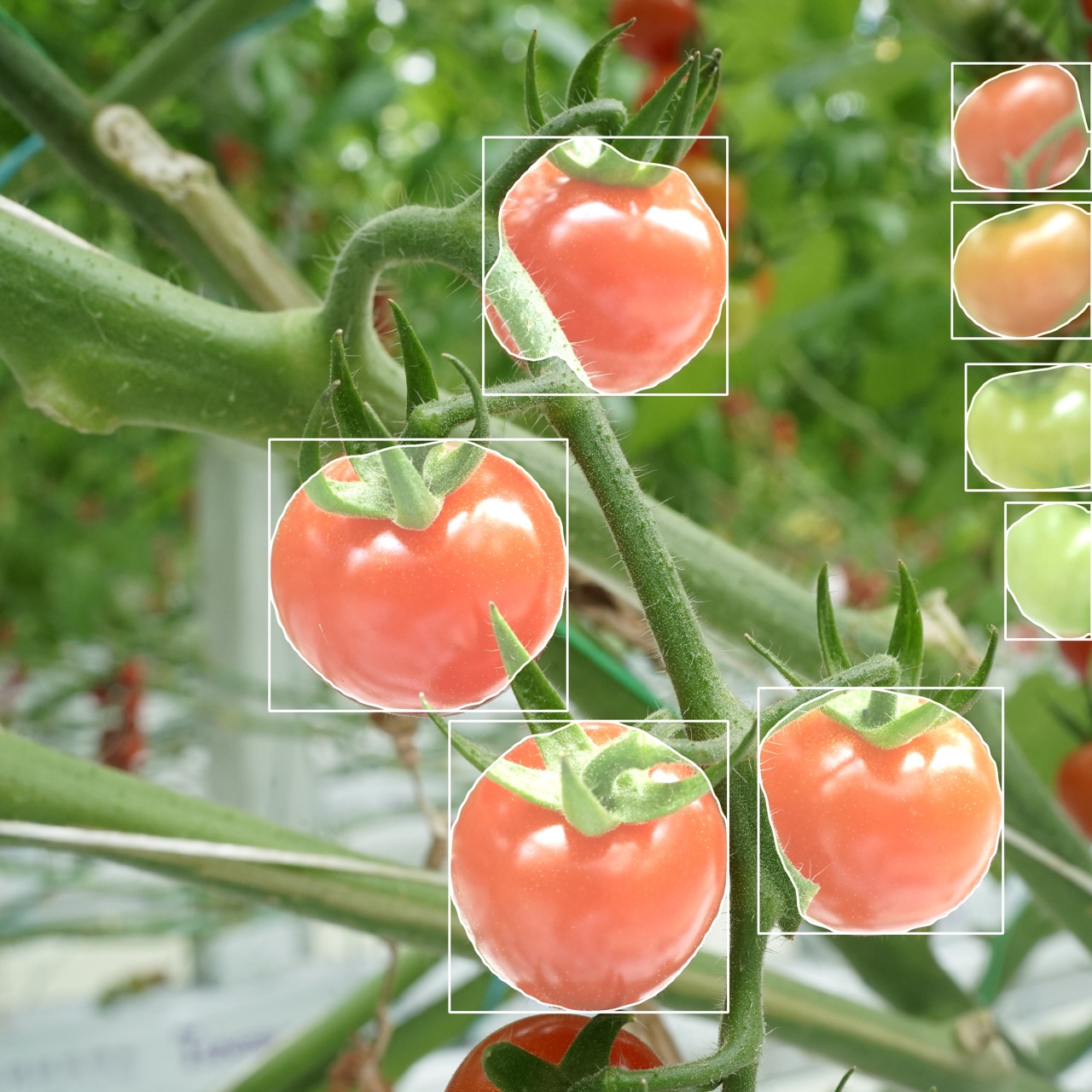}
     \caption{YOLACT}
   \end{subfigure}
   \hfill
   \begin{subfigure}{0.24\linewidth}
    \includegraphics[width=1.0\linewidth]{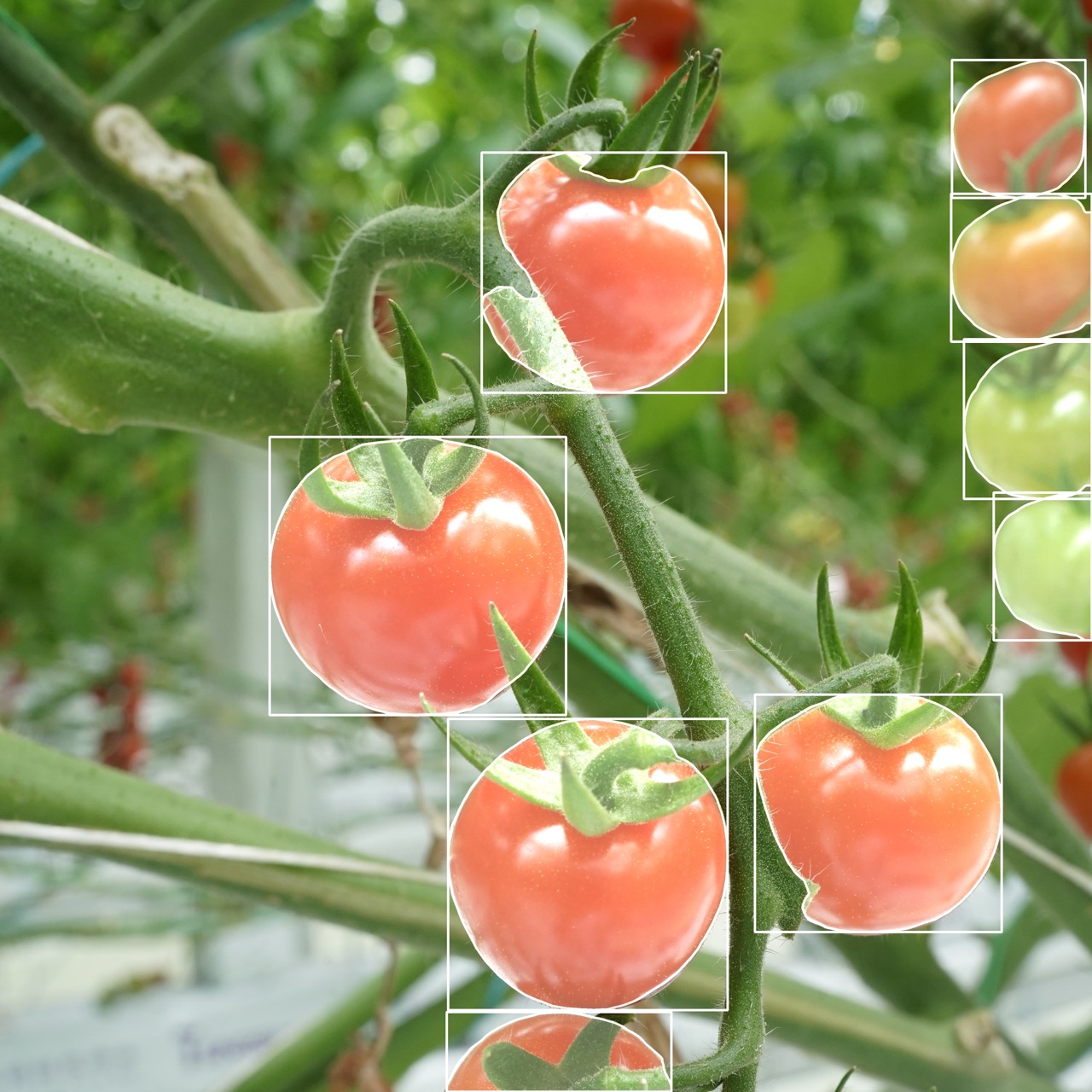}
     \caption{RTMDet}
   \end{subfigure}
 
   \begin{subfigure}{0.24\linewidth}
    \includegraphics[width=1.0\linewidth]{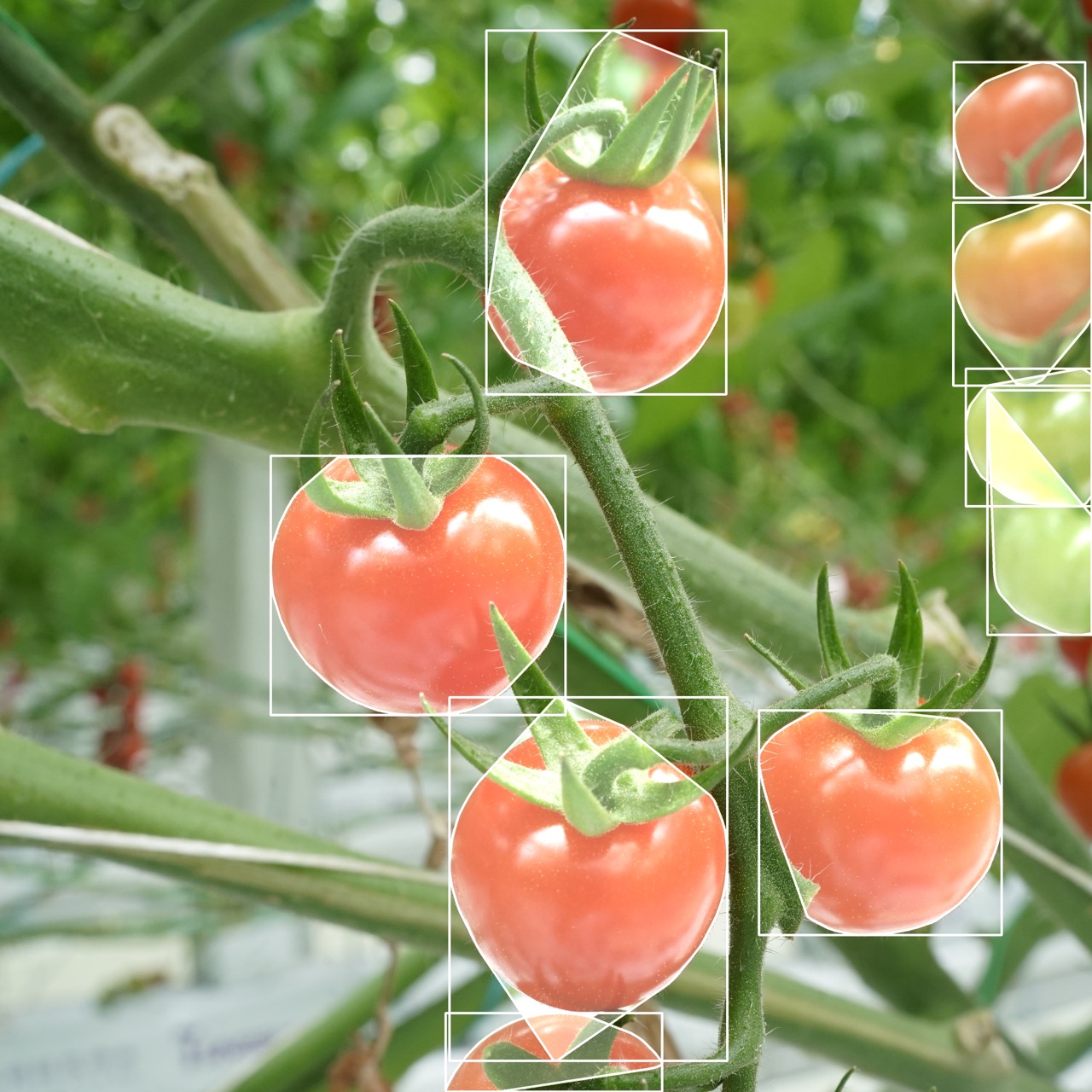}
     \caption{SAM2}
   \end{subfigure}
   \hfill
   \begin{subfigure}{0.24\linewidth}
    \includegraphics[width=1.0\linewidth]{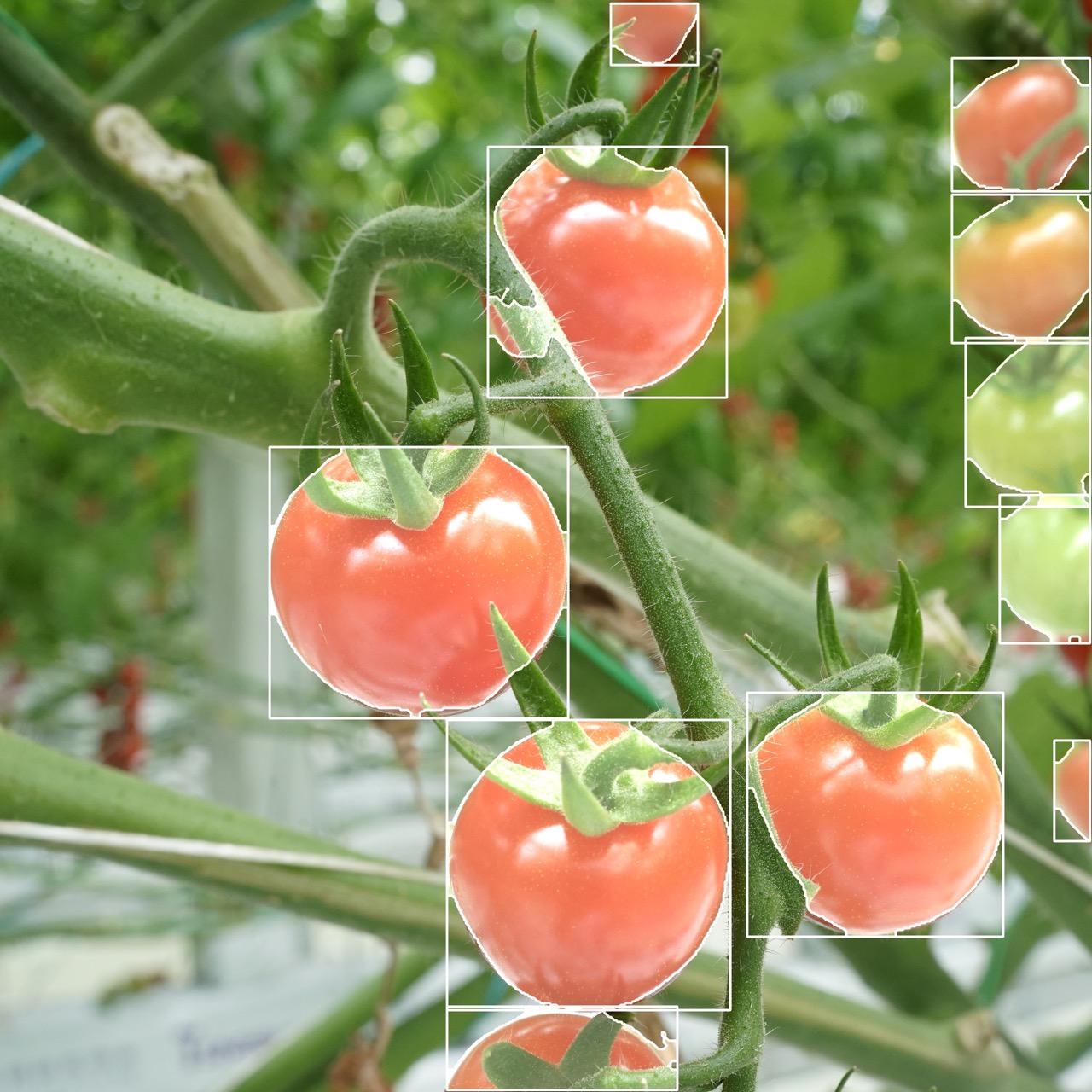}
     \caption{YOLO11n-seg}
   \end{subfigure}
   \hfill
   \begin{subfigure}{0.24\linewidth}
    \includegraphics[width=1.0\linewidth]{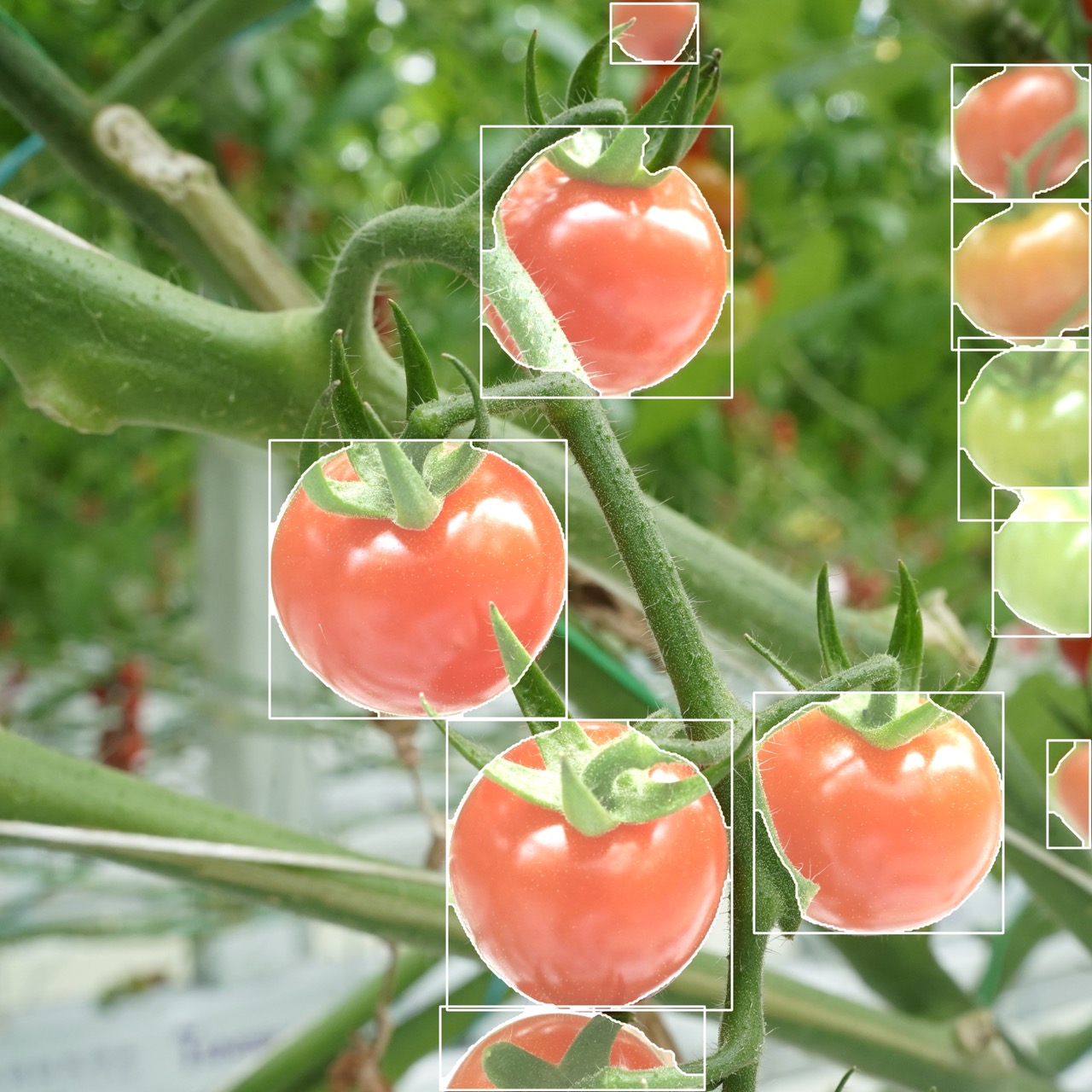}
     \caption{YOLO11x-seg}
   \end{subfigure}
   \hfill
   \begin{subfigure}{0.24\linewidth}
    \includegraphics[width=1.0\linewidth]{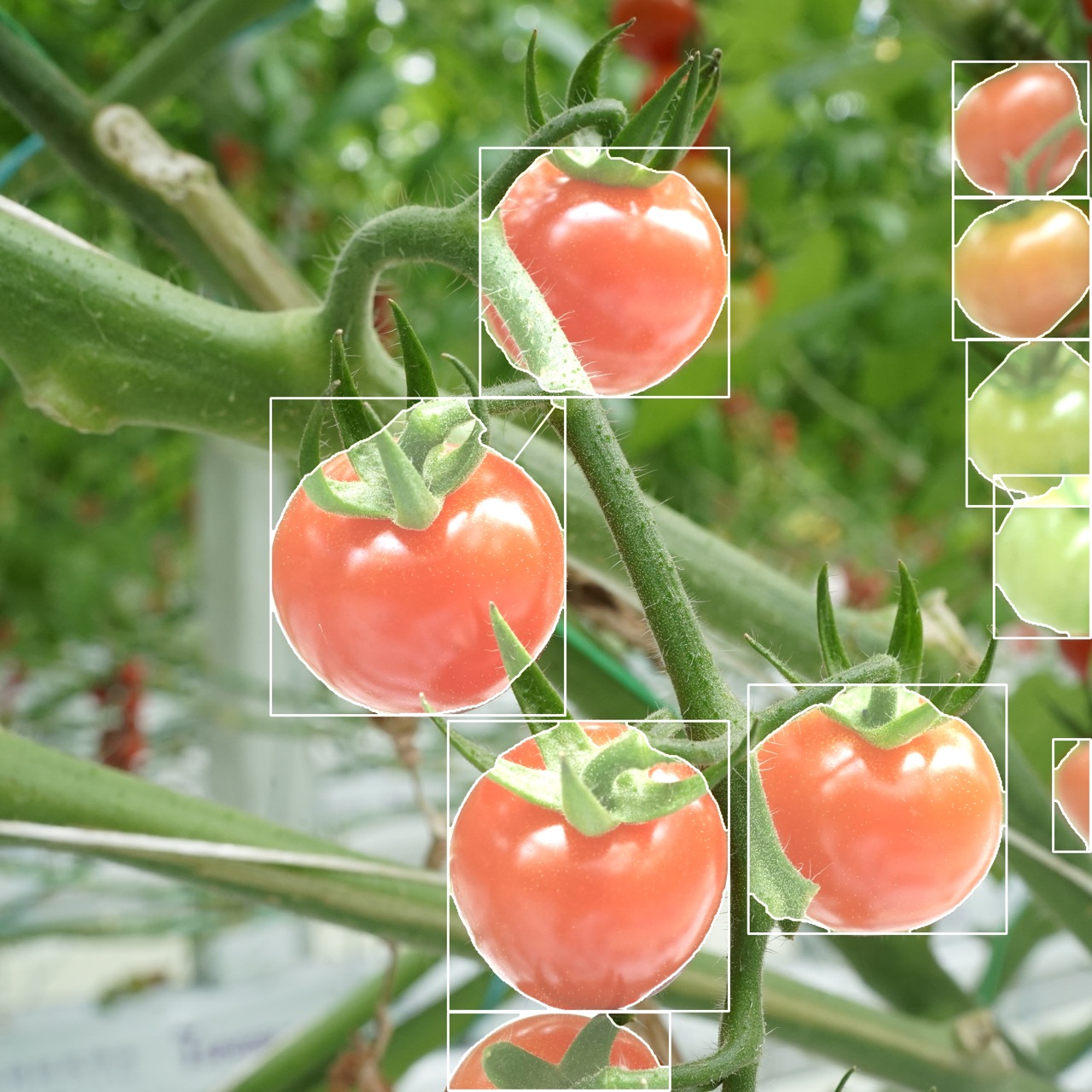}
     \caption{Ours}
   \end{subfigure}

  \begin{subfigure}{0.24\linewidth}
    \includegraphics[width=1.0\linewidth]{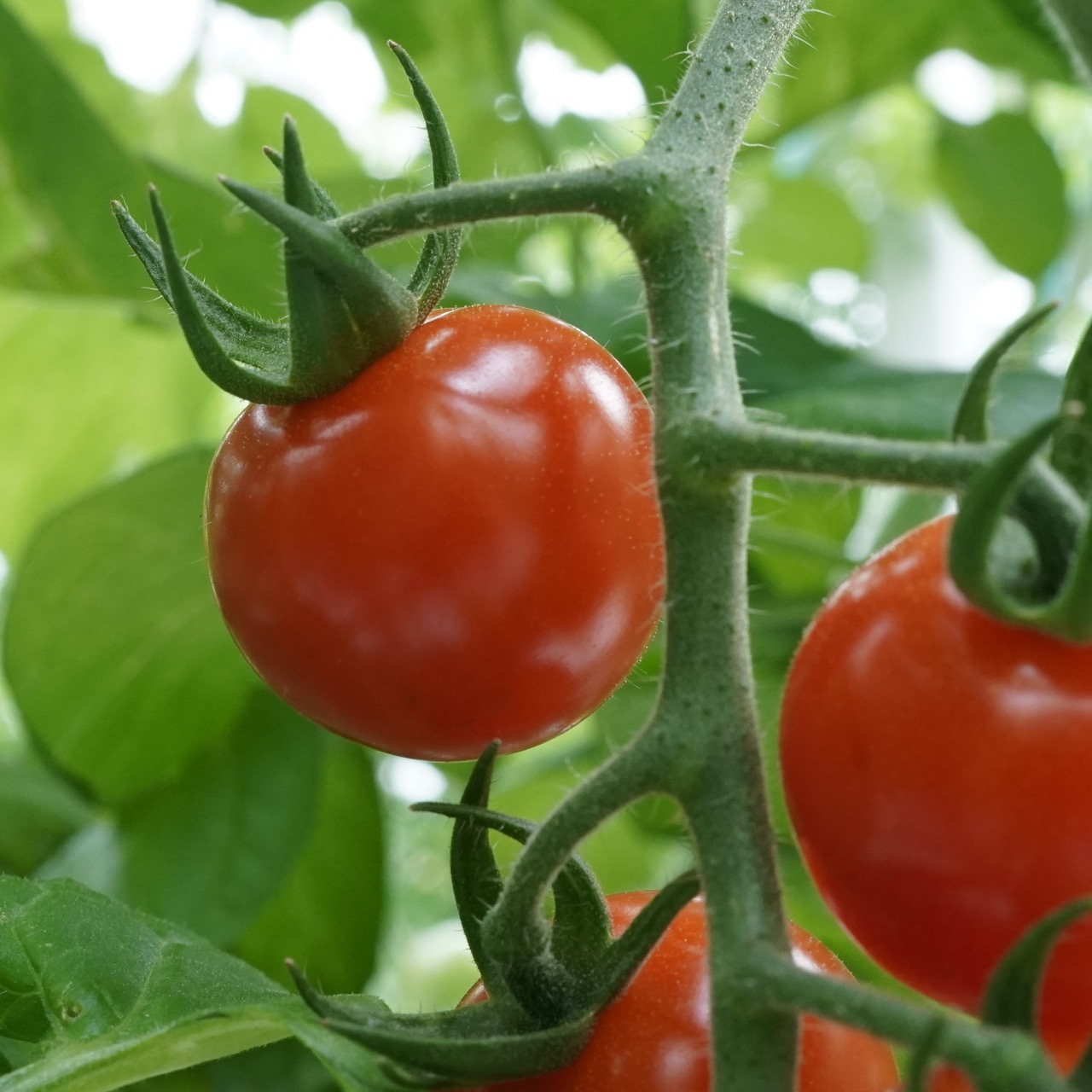}
     \caption{Input NO.3}
   \end{subfigure}
   \hfill
   \begin{subfigure}{0.24\linewidth}
    \includegraphics[width=1.0\linewidth]{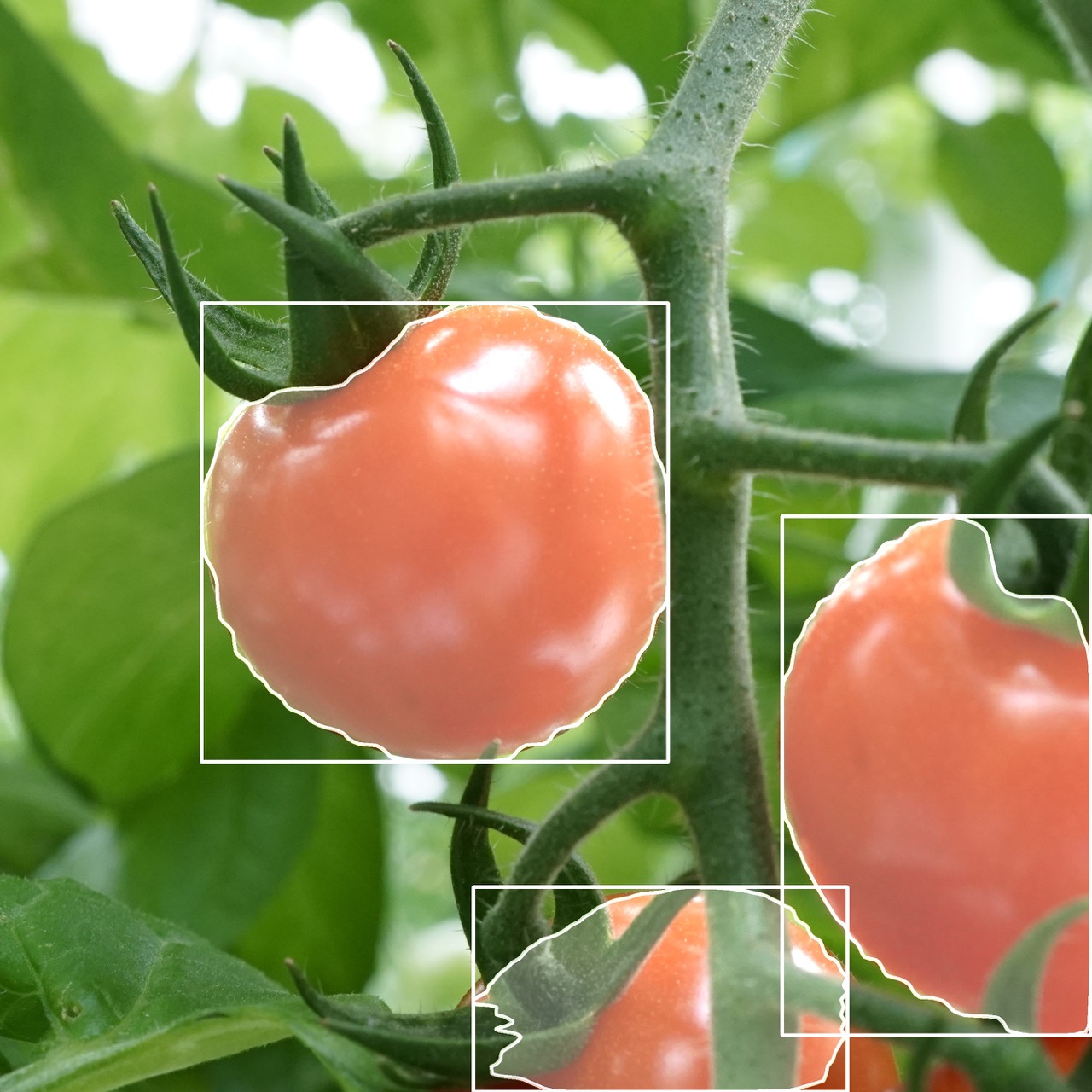}
     \caption{Mask R-CNN}
   \end{subfigure}
   \hfill
   \begin{subfigure}{0.24\linewidth}
    \includegraphics[width=1.0\linewidth]{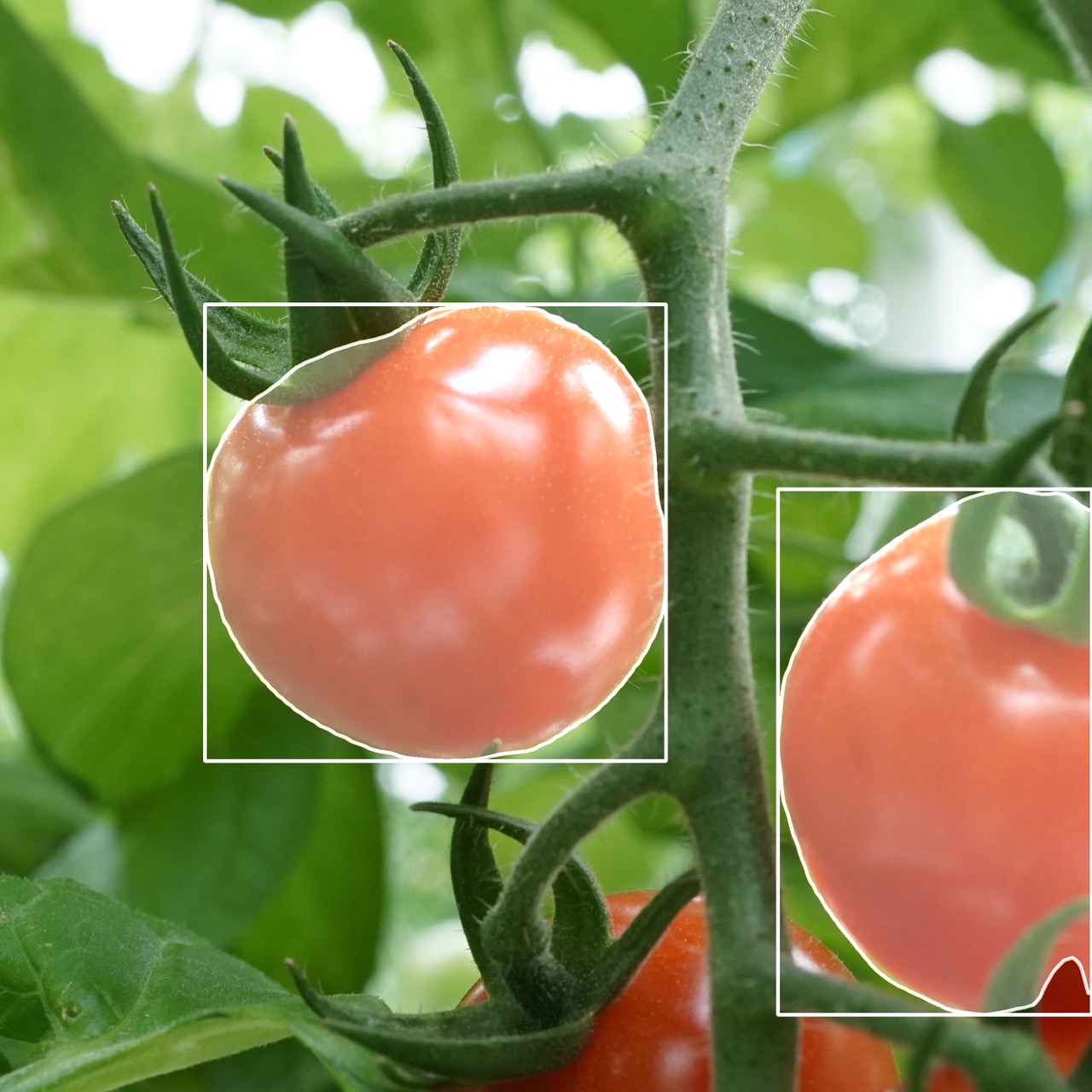}
     \caption{YOLACT}
   \end{subfigure}
   \hfill
   \begin{subfigure}{0.24\linewidth}
    \includegraphics[width=1.0\linewidth]{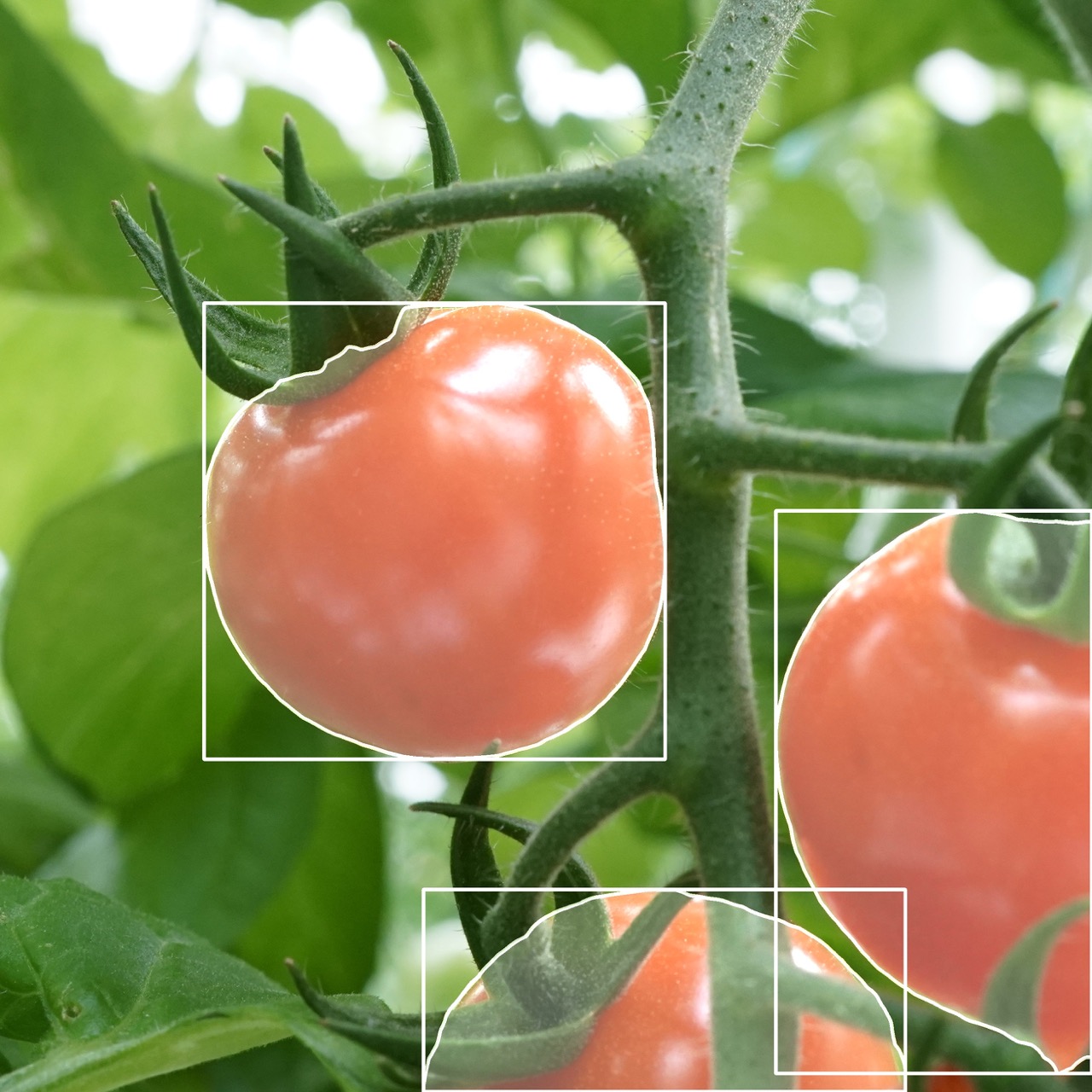}
     \caption{RTMDet}
   \end{subfigure}

  \begin{subfigure}{0.24\linewidth}
    \includegraphics[width=1.0\linewidth]{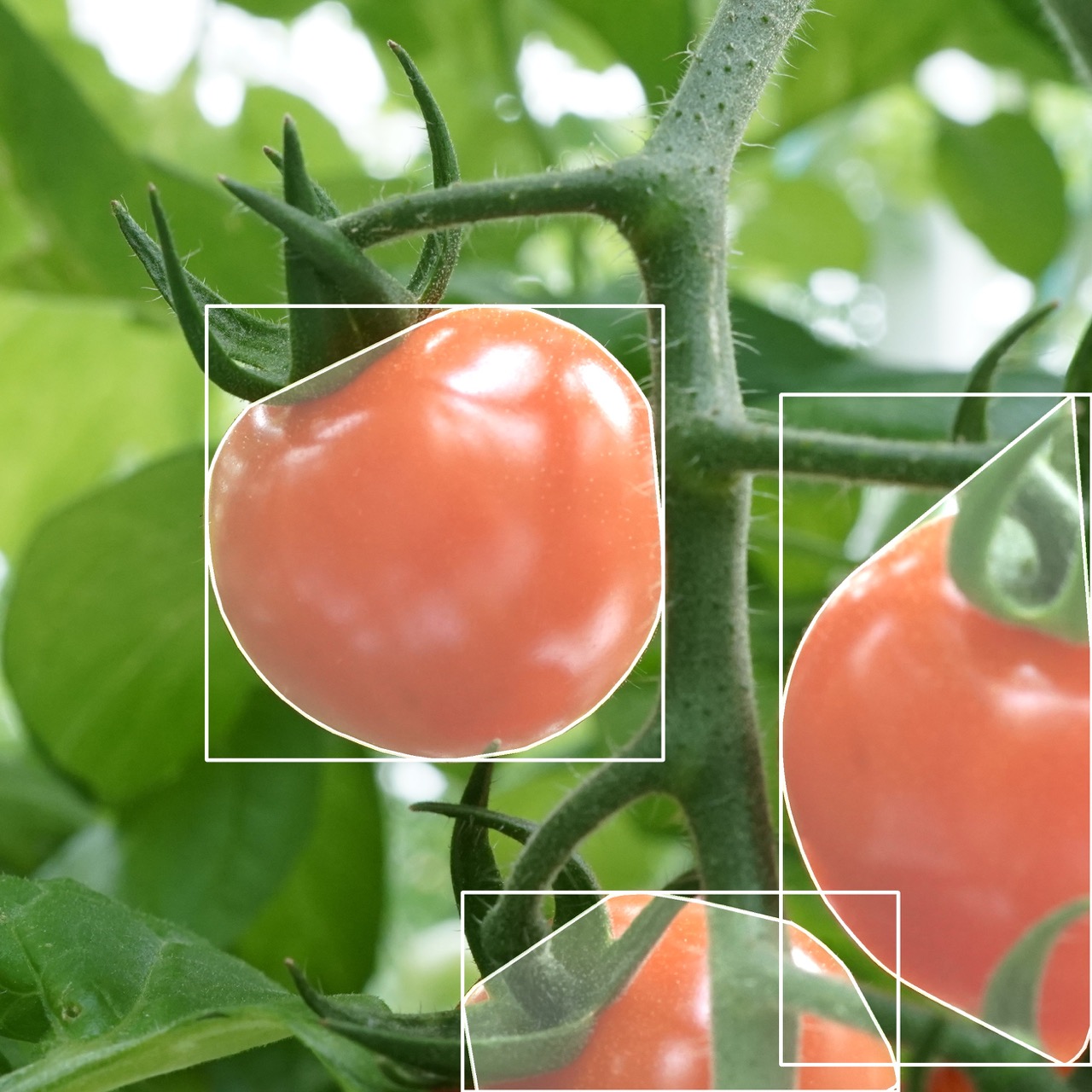}
     \caption{SAM2}
   \end{subfigure}
   \hfill
   \begin{subfigure}{0.24\linewidth}
    \includegraphics[width=1.0\linewidth]{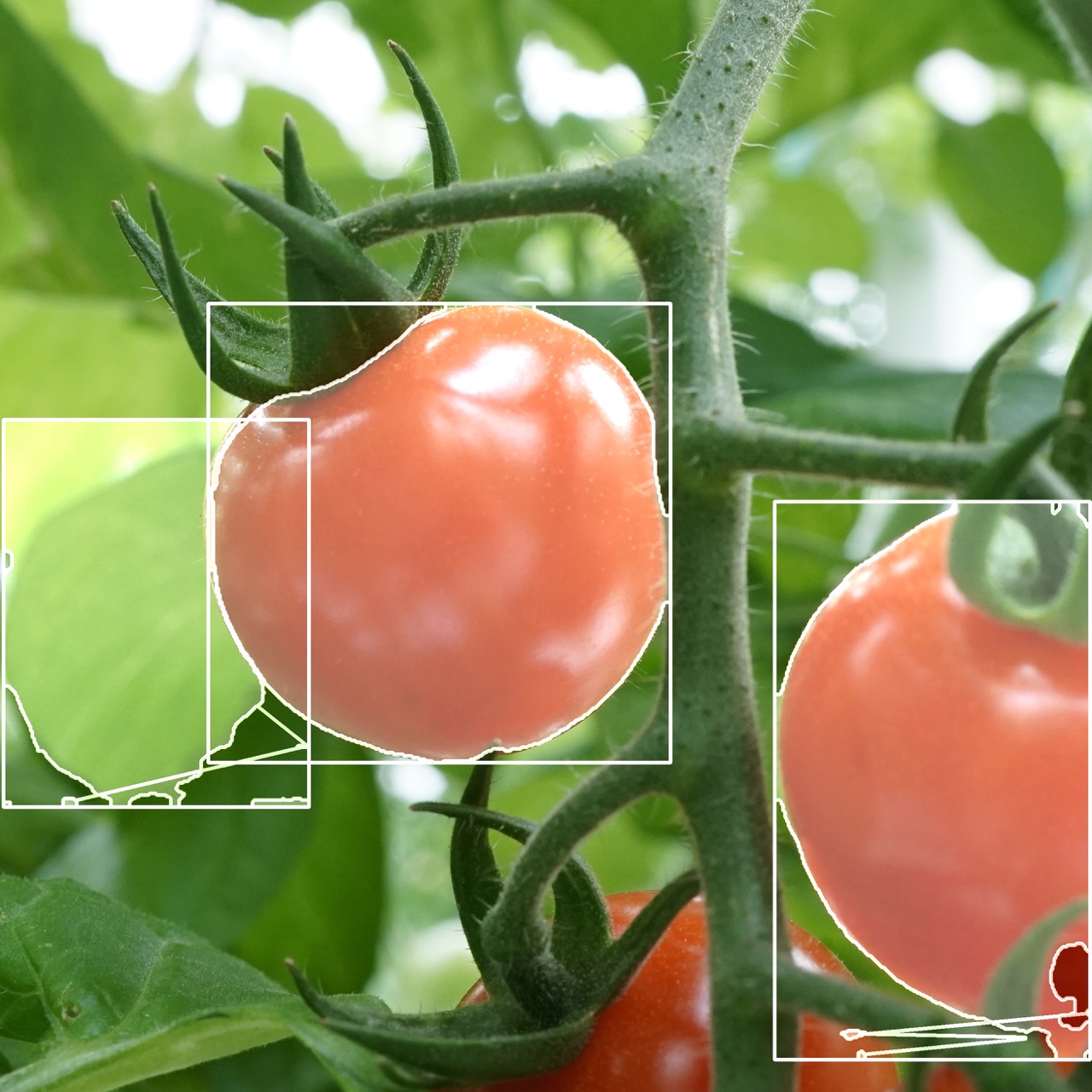}
     \caption{YOLO11n-seg}
   \end{subfigure}
   \hfill
   \begin{subfigure}{0.24\linewidth}
    \includegraphics[width=1.0\linewidth]{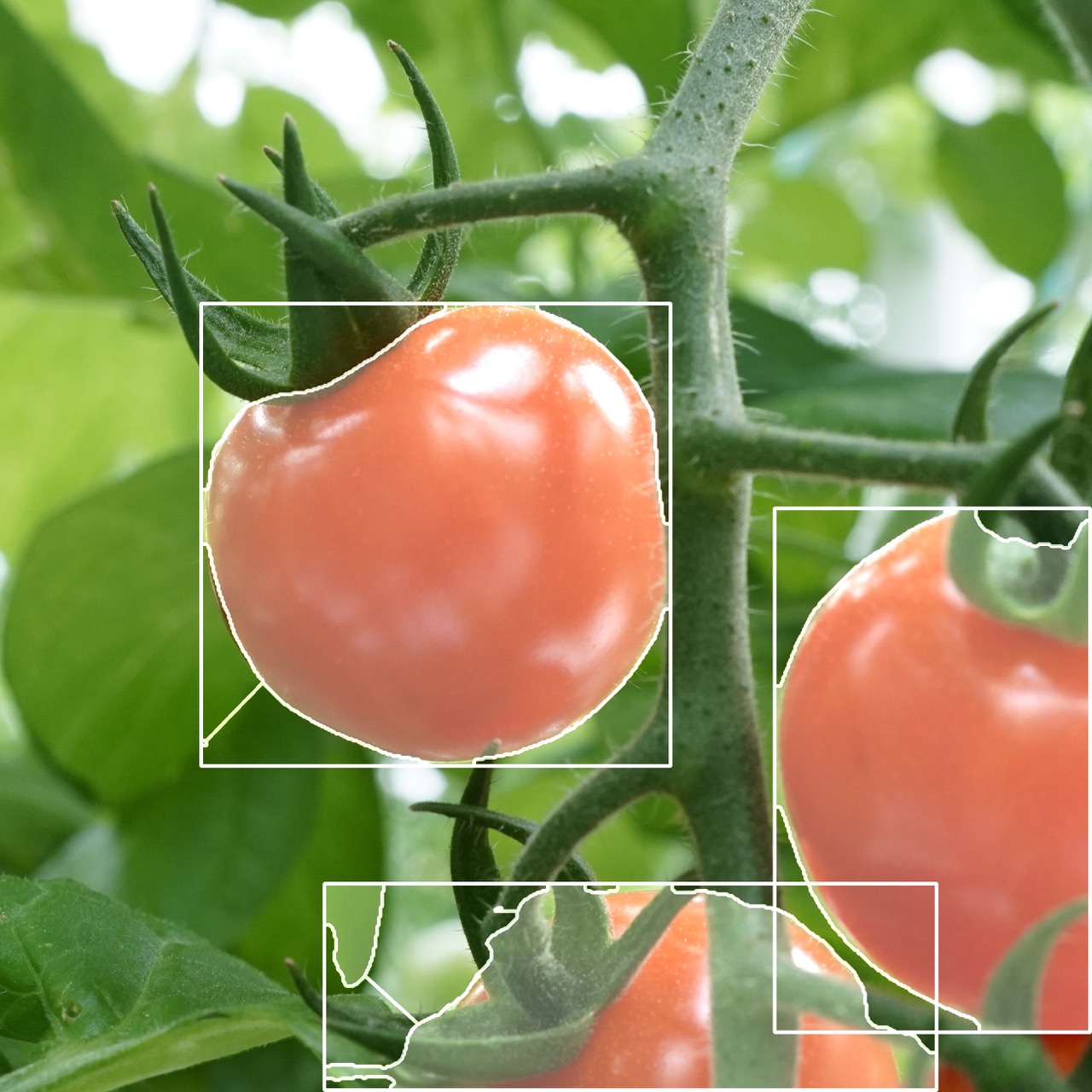}
     \caption{YOLO11x-seg}
   \end{subfigure}
   \hfill
   \begin{subfigure}{0.24\linewidth}
    \includegraphics[width=1.0\linewidth]{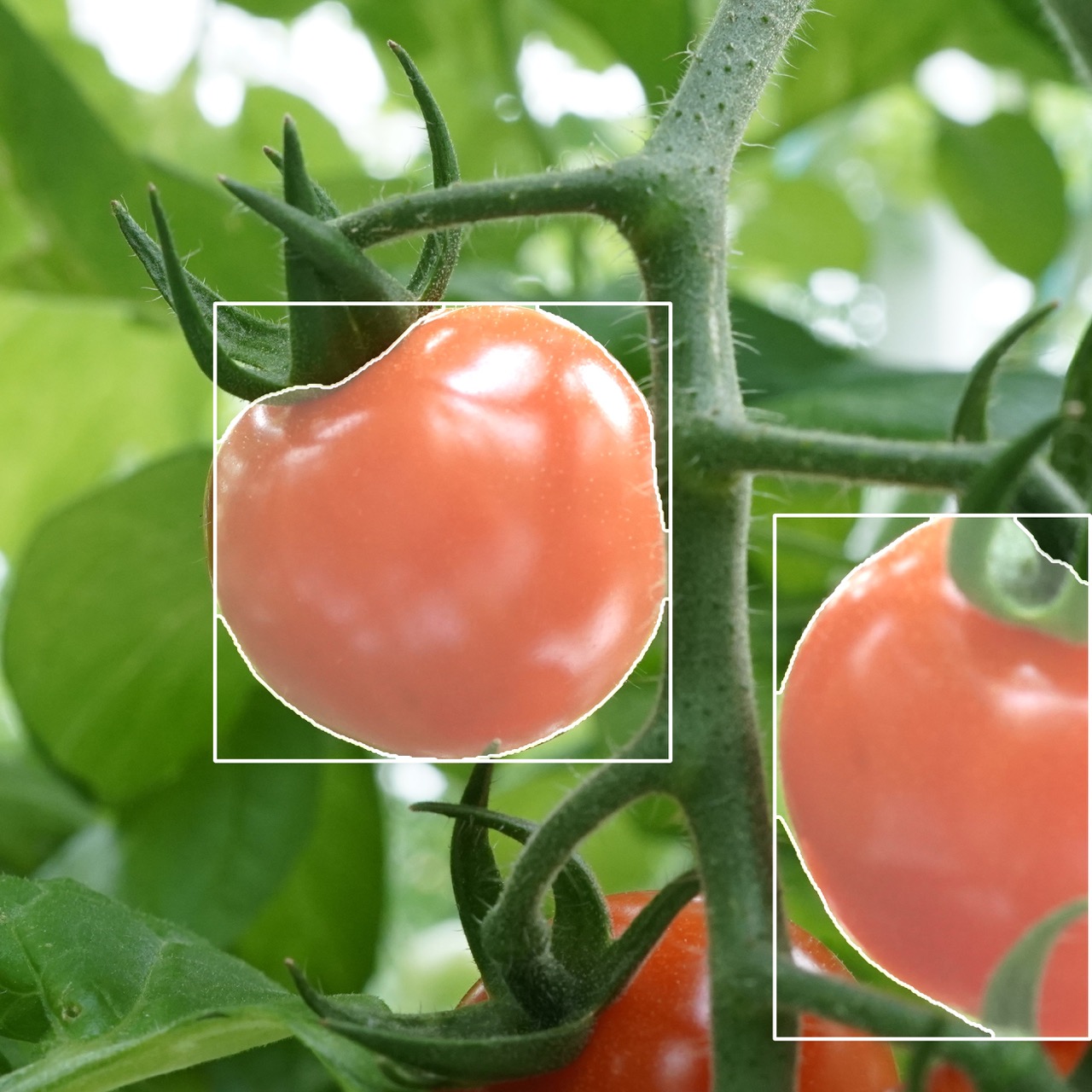}
     \caption{Ours}
   \end{subfigure}

  \caption{Instance segmentation model comparison demonstration. (Zoom in for better observation)}
  \label{fig:instance_segmentation_model_comparison_demonstration}
\end{figure}

\begin{table*}[t]
  \centering
  \resizebox{\textwidth}{!}{ 
  \begin{tabular}{@{}cccc|cc|ccc|cl@{}}
    \toprule
      & EdgeAttention & EdgeLoss & EdgeBoost & Weights & FPS & P & R & mAP50 & mEE \\
    \midrule
baseline &      &         &         & 45.2 M & 76.220 & 0.943 & 0.940 & 0.970 & 5.641 \% \\
      & $\surd$ &         &         & 48.7 M & 78.247 & 0.972 & 0.907 & 0.966 & 4.269 \% \\
      &         & $\surd$ &         & 45.2 M & 75.873 & 0.940 & 0.927 & 0.966 & 5.306 \% \\
      &         &         & $\surd$ & 45.2 M & 78.989 & 0.924 & 0.954 & 0.969 & 5.232 \% \\
      & $\surd$ & $\surd$ &         & 48.7 M & 78.003 & 0.928 & 0.940 & 0.961 & 2.948 \% \\
      & $\surd$ &         & $\surd$ & 48.7 M & 78.370 & 0.976 & 0.901 & 0.960 & 3.989 \% \\
      &         & $\surd$ & $\surd$ & 45.2 M & 80.000 & 0.941 & 0.934 & 0.955 & 3.082 \% \\
 Ours & $\surd$ & $\surd$ & $\surd$ & 48.7 M & 76.336 & 0.986 & 0.906 & 0.955 & 2.963 \% \\
    \bottomrule
  \end{tabular}
  }
  \caption{Ablation experiment results: The baseline refers to the raw YOLO11m-seg without any improvements. 
  Every check mark represents that corresponding module is added to YOLO11m-seg. 
  Cyan or orange background represent that corresponding score is better or worse than baseline, respectively. 
  The darker the background, the larger the change. 
  }
  \label{tab:ablation_table}
\end{table*}

\par Automation degree: ZG must put the calibration next to the tomato fruit. 
TZ must place three cameras at different specific perspectives. 
Our method, however, does not ask for any conditions of RGB image. 

\par Computation burden: TZ's 3D reconstruction and Ours' instance segmentation and depth estimation are higher than ZG's detection on computation burden, 
which may limit the deployment prospect. 

\par In summary, TomatoScanner achieves full automation and greatly reduces input complexity, meanwhile, reaches the equally excellent phenotyping accuracy. 

\subsubsection{Instance Segmentation Models}

\par We compare EdgeYOLO with five state-of-the-art instance segmentation models \cite{MaskR-CNN,YOLACT,RTMDet,SAM2,YOLO11}. 
All models are trained on the same dataset with same software and hardware (except for SAM2 which does not need training). 

\par The comparison results are demonstrated in \cref{tab:instance_segmentation_model_comparison_table} and \cref{fig:instance_segmentation_model_comparison_demonstration}. 
As a classical instance segmentation model, Mask R-CNN achieves excellent balance between speed and accuracy, but the 334.8 weights size limit the downstream deployment prospect. 
Similarly, RTMDet is too heavy and too slow to deal with real-time task. 
YOLACT achieves acceptable lightweight, but its mAP50 and mEE are only 0.923 and 6.149 \%, respectively, which demonstrates the terrible segmentation accuracy on all portions. 
YOLO11n-seg is extremely lightweight with only 6 M weights size, meanwhile, reaches the best mAP 50 of 0.969. 
YOLO11x-seg has more than 20 times the weights size of YOLO11n-seg, but achieves inferior accuracy. 
The possible reason is that tomato fruit is not a very difficult classification to be segmented, the overly complex architecture may have negative impact on the feature extraction. 
Compared to YOLO11n-seg which segment better on entirety, our EdgeYOLO segment better on edge portion. 

\par In summary, every model processes different advantages. 
For tomato phenotyping task, however, the focus-on-edge characteristic makes EdgeYOLO the best option. 

\subsection{Ablation Experiment}

\par EdgeYOLO, our proposed instance segmentation model, plays the core role in TomatoScanner.
To improve the segmentation accuracy on edge portion, three new modules - EdgeAttention, EdgeLoss and EdgeBoost - are proposed and added to the raw YOLO11m-seg. 
We implement ablation experiment to verify the effectiveness and impact of each module. 

\begin{figure}[b]
  \centering
  \includegraphics[width=1.0\linewidth]{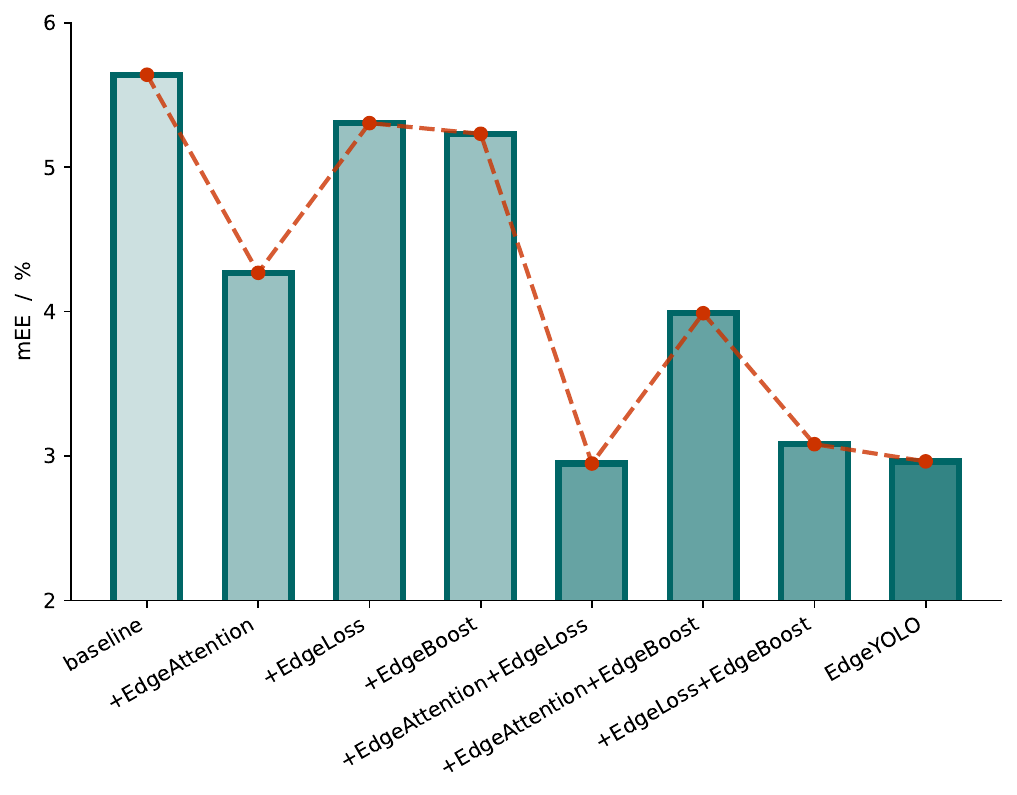}
  \caption{Ablation experiment statistic plot of mEE  
  }
  \label{fig:ablation_plot}
\end{figure}

\par The ablation experiment results are shown in \cref{tab:ablation_table} and \cref{fig:ablation_plot}. 
When EdgeAttention, EdgeLoss or EdgeBoost is added solely, mEE reduces 1.372, 0.335, 0.409 percent point, respectively, 
proving that every module can improve the edge segmentation accuracy. 
Meanwhile, the tiny sacrifice of mAP50 is acceptable. 
When any two of the three modules are added, mEE reduces furthermore, 
which demonstrates that the three modules do not cause negative interference with each other. 
When all modules are added, P reaches the best score and R reaches the lowest score. 
In our tomato phenotypic measurement task, 
it is acceptable to miss a goal but unacceptable to misjudge, 
therefore, precision is much more important than recall. 
Thus, the preference between outstanding improvement of P \& mEE and acceptable sacrifice of R \& mAP50 are the most suitable. 
In addition, weights size and FPS keep splendid score whatever modules are added. 

\par In summary, each module makes effective impact on different aspects. 
If and only if all three modules are added, EdgeYOLO achieves the best performance.

\section{Conclusion}
\label{sec:conclusion}

\par In tomato greenhouse, phenotypic measurement is meaningful for researchers and farmers to monitor crop growth, thereby precisely control environmental conditions in time, leading to better quality and higher yield. 
Traditional phenotyping mainly relies on manual measurement, which is accurate but inefficient, more importantly, endangering the health and safety of people. 
Several studies have explored computer vision-based methods to replace manual phenotyping. 
However, the 2D-based need extra calibration, or cause destruction to fruit, or can only measure limited and meaningless traits. 
The 3D-based need extra depth camera, which is expensive and unacceptable for most farmers. 
Thus, developing a more advanced 2D-based method with excellent phenotyping accuracy is meaningful and urgent. 

\par In this paper, we propose a non-contact tomato fruit phenotyping method, titled TomatoScanner, 
where RGB image is all you need for input. 
We establish self-built Tomato Phenotype Dataset to test TomatoScanner, which achieves excellent phenotyping on width, height, vertical area and volume, 
with median relative error of 5.63\%, 7.03\%, -0.64\% and 37.06\%, respectively. 
EdgeYOLO is the core instance segmentation model in TomatoScanner. 
We propose and add three innovative modules - EdgeAttention, EdgeLoss and EdgeBoost - into EdgeYOLO, to enhance the segmentation accuracy on edge portion. 
Precision and mean Edge Error greatly improve from 0.943 and 5.641\% to 0.986 and 2.963\%, respectively. 
Meanwhile, EdgeYOLO keeps lightweight and efficient, with 48.7 M weights size and 76.34 FPS. 

\par During the test experiment, we also find two main weaknesses on TomatoScanner: 
1. Calculation logic of width and height are similar, but accuracy of height is relatively worse. 
Specifically, predicted height is usually bigger than groung truth height. 
The possible reason is that in some images of fruit, the capture angles are not appropriate, in which carpopodiums are not on the right above, leading to the tilted poses of fruit. 
Therefore, the predicted heights are not the actual height. 
2. The accuracy of volume is inferior. 
The possible reason is that volume is calculated by the integral of horizontal crossing area to vertical axis but not direct extraction, 
where multi parameters may introdece errors. 
In the final, the cumulation of errors from level to level leads to the untimate great error. 

\par In the future, in addition to the fixing and improvement of above weaknesses, 
we plan to deploy TomatoScanner on intelligent robotic dog, 
which can self-drive in the greenhouse, 
meanwhile, capture and send RGB image of tomato fruit to TomatoScanner for data processing, 
to accomplish the complete phenotyping solution. 

{
    \small
    \bibliographystyle{unsrt}
    \bibliography{main}
}
\newpage
\appendix
\onecolumn
\setcounter{table}{0}   
\setcounter{figure}{0}
\setcounter{section}{0}
\renewcommand{\thetable}{A\arabic{table}}
\renewcommand{\thefigure}{A\arabic{figure}}
\renewcommand{\thesection}{A\arabic{section}}

\section{Training Related Appendix}
\begin{figure*}[h]
   \centering
      \includegraphics[width=1.0\linewidth]{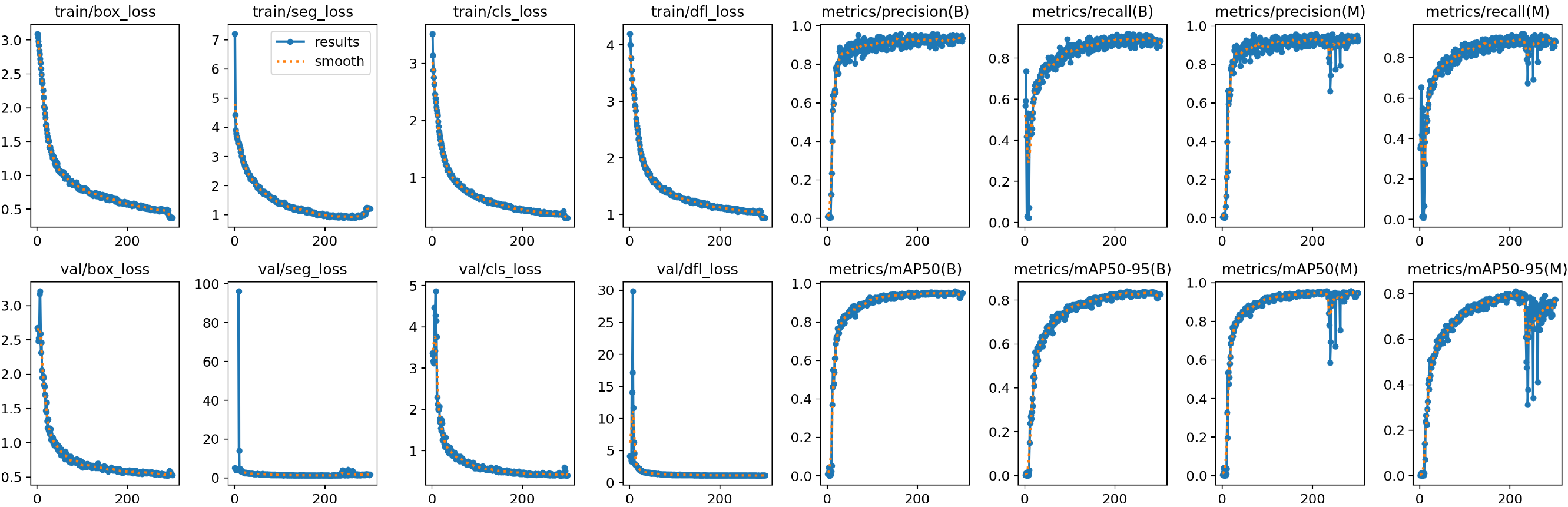}
      \caption{EdgeYOLO segmentor training process illustration}
      \label{fig:EdgeYOLO_training}
   \end{figure*}

\begin{figure*}[h]
\centering
      \includegraphics[width=1.0\linewidth]{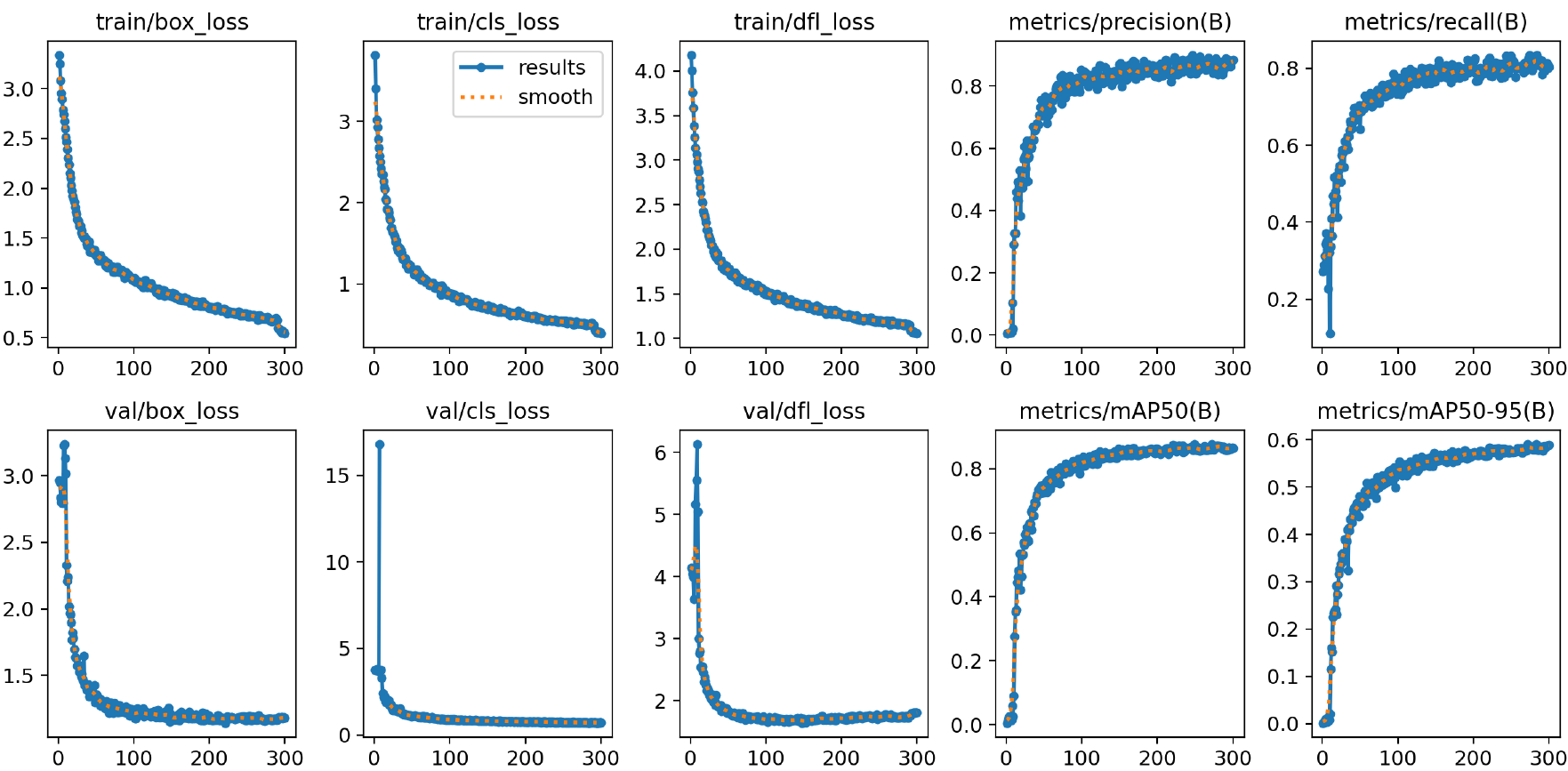}
      \caption{YOLO11 detector training process illustration}
      \label{fig:YOLO11_training}
\end{figure*}

\newpage
\section{Test Experiment Related Appendix}

\begin{table*}[h]
   \centering
   \resizebox{\textwidth}{!}{ 
   \begin{tabular}{@{}cc|rrr|rrr|rrr|rrrl@{}}
      \toprule
      Plant & Fruit & $W_{t}$ & $W_{p}$ & $W_{e}$ & $H_{t}$ & $H_{p}$ & $H_{e}$ & $A_{t}$ & $A_{p}$ & $A_{e}$ & $V_{t}$ & $V_{p}$ & $V_{e}$ \\
      \midrule
      1 & 1 & 2.821 & 2.915 & +3.328 & 2.653 & 2.857 & +7.682 & 6.75 & 6.533 & -3.216 & 11 & 12.804 & +16.398 \\
      \midrule
      2 & 1 & 2.624 & 2.569 & -2.103  & 2.309 & 2.237 & -3.102  & 4.50 & 4.540 & +0.895  & 7 & 7.850 & +12.142 \\
        & 2 & 2.608 & 2.385 & -8.563  & 2.308 & 2.081 & -9.842  & 5.25 & 4.019 & -23.450 & 7 & 6.469 & -7.583 \\
      \midrule
      6 & 1 & 2.288 & 2.393 & +4.573 & 2.094 & 2.171 & +3.691 & 4.25 & 4.190 & -1.416 & 4 & 6.807 & +70.185 \\
      \midrule
      7 & 1 & 2.610 & 2.864 & +9.739 & 2.446 & 2.484 & +1.568 & 5.50 & 5.706 & +3.742 & 8 & 10.965 & +37.064 \\
      \midrule
      10 & 1 & 2.459 & 2.544 & +3.465 & 2.301 & 2.323 & +0.955 & 5.25 & 4.737 & -9.771 & 5 & 8.137 & +62.747 \\
      \midrule
      11 & 1 & 2.563 & 2.791 & +8.887 & 2.448 & 2.570 & +4.972 & 5.00 & 5.835 & +16.701 & 9 & 10.994 & +22.155 \\
      \midrule
      12 & 1 & 2.754 & 3.056 & +10.958 & 2.566 & 2.950 & +14.949 & 5.50 & 7.166 & +30.284 & 12 & 14.722 & +22.687 \\
      \midrule
      13 & 1 & 2.658 & 2.970 & +11.741 & 2.388 & 2.681 & +12.289 & 5.75 & 6.346 & +10.369 & 9 & 12.774 & +41.930 \\
      \midrule
      14 & 1 & 2.579 & 2.903 & +12.564 & 2.318 & 2.534 & +9.307 & 5.50 & 6.118 & +11.235 & 7 & 12.258 & +75.109 \\
      \midrule
      15 & 1 & 3.297 & 3.298 & +0.045 & 2.929 & 2.901 & -0.945 & 9.50 & 7.620 & -19.791 & 15 & 16.887 & +12.578 \\
      \midrule
      16 & 1 & 2.863 & 2.965 & +3.568 & 2.418 & 2.588 & +7.029 & 6.75 & 6.042 & -10.485 & 10 & 12.083 & +20.828 \\
      \midrule
      17 & 1 & 2.724 & 2.798 & +2.730 & 2.464 & 2.531 & +2.735 & 6.50 & 5.643 & -13.185 & 9 & 10.677 & +18.635 \\
      \midrule
      18 & 1 & 2.461 & 2.224 & -9.631 & 2.140 & 1.916 & -10.473 & 4.75 & 3.453 & -27.296 & 6 & 5.211 & -13.149 \\
      \midrule
      19 & 1 & 2.392 & 2.712 & +13.391 & 2.073 & 2.391 & +15.340 & 4.50 & 5.275 & +17.214 & 6 & 9.724 & +62.060 \\
      \midrule
      20 & 1 & 2.560 & 2.521 & -1.512 & 2.354 & 2.320 & -1.462 & 5.50 & 4.688 & -14.761 & 9 & 7.977 & -11.369 \\
      \midrule
      21 & 1 & 2.714 & 2.867 & +5.633 & 2.462 & 2.712 & +10.140 & 6.50 & 6.180 & -4.928 & 9 & 11.951 & +32.794 \\
      \midrule
      22 & 1 & 2.592 & 2.757 & +6.384 & 2.319 & 2.476 & +6.784 & 6.25 & 5.301 & -15.192 & 7 & 9.689 & +38.420 \\
      \midrule
      23 & 1 & 2.680 & 2.792 & +4.194 & 2.394 & 2.518 & +5.162 & 7.00 & 5.681 & -18.841 & 9 & 10.745 & +19.389 \\
      \midrule
      24 & 1 & 2.640 & 2.768 & +4.847 & 2.468 & 2.389 & -3.198 & 6.50 & 5.290 & -18.622 & 8 & 9.835 & +22.941 \\
      \midrule
      25 & 1 & 2.578 & 2.647 & +2.693 & 2.254 & 2.531 & +12.278 & 5.75 & 5.333 & -7.246 & 6 & 9.538 & +58.963 \\
      \midrule
      26 & 1 & 2.902 & 3.141 & +8.240 & 2.614 & 2.913 & +11.436 & 7.25 & 7.204 & -0.639 & 11 & 15.155 & +37.775 \\
      \midrule
      28 & 1 & 3.038 & 2.971 & -2.197 & 2.751 & 2.663 & -3.215 & 8.25 & 6.355 & -22.973 & 13 & 12.751 & -1.913 \\
      \midrule
      29 & 1 & 2.892 & 3.073 & +6.272 & 2.613 & 2.975 & +13.870 & 6.75 & 6.965 & +3.191 & 10 & 13.932 & +39.316 \\
      \midrule
      30 & 1 & 2.698 & 3.117 & +15.536 & 2.548 & 2.825 & +10.883 & 6.50 & 7.064 & +8.678  & 10 & 14.803 & +48.025 \\
         & 2 & 2.689 & 2.977 & +10.713 & 2.598 & 2.884 & +10.996 & 6.75 & 6.855 & +1.562  & 9 & 13.759  & +52.876 \\
         & 3 & 2.601 & 2.814 &   +8.175  & 2.509 & 2.837 & +13.072 & 5.50 & 6.352 & +15.491 & 8 & 12.102  & +51.272 \\
         & 4 & 2.422 & 2.732 &   +12.795 & 2.296 & 2.592 & +12.884 & 5.25 & 5.920 & +12.764 & 7 & 11.206  & +60.081 \\
         & 5 & 2.329 & 2.452 &   +5.269  & 2.128 & 2.463 & +15.760 & 4.00 & 4.698 & +17.444 & 6 & 7.685   & +28.084 \\
         & 6 & 1.992 & 2.172 &   +9.012  & 1.889 & 2.187 & +15.780 & 3.50 & 3.699 & +5.697  & 3 & 5.374   & +79.129 \\
      \midrule
      31 & 1 & 3.338 & 3.734 & +11.864 & 3.191 & 3.362 & +5.362 & 10.25 & 10.364 & +1.108 & 19 & 26.551 & +39.740 \\
     \bottomrule
   \end{tabular}}
   \caption{Complete test experiment results: 
   W, H, A and V refer to width, height, vertical area and volume, respectively. 
   Subscript t, p and e refer to ground truth value, predicted value and relative error, respectively. 
   Positive or negative sign represent that predicted value is bigger or smaller than ground truth value, respectively. 
   }
   \label{tab:complete_test_table}
\end{table*}

\end{document}